\documentclass[letterpaper]{article}
\usepackage{aaai}
\usepackage{times}
\usepackage{helvet}
\usepackage{courier}
\usepackage{subcaption}
\usepackage{multirow}
\usepackage{comment}
\usepackage{adjustbox}
\usepackage{lipsum}
\usepackage{mathrsfs} 
\usepackage{amsmath,amssymb,amsfonts,graphicx,float}

\usepackage{tikz} 
 \usepackage{pgfplots}
\pgfplotsset{compat=newest}
\usetikzlibrary{plotmarks}
\usetikzlibrary{arrows.meta}
\usepgfplotslibrary{patchplots}
\usetikzlibrary{arrows,decorations.markings}
\usetikzlibrary{decorations.pathreplacing}
\usetikzlibrary{automata,arrows,positioning,calc}

\begin{document}
%
\title{Deceptive Decision-Making Under Uncertainty}
\author{Yagiz Savas, Christos K. Verginis, Ufuk Topcu}
\maketitle
\begin{abstract}
\begin{quote}
We study the design of autonomous agents that are capable of deceiving outside observers about their intentions while carrying out tasks in stochastic, complex environments. By modeling the agent's behavior as a Markov decision process, we consider a setting where the agent aims to reach one of multiple potential goals while deceiving outside observers about its true goal. We propose a novel approach to model observer predictions based on the principle of maximum entropy and to efficiently generate deceptive strategies via linear programming. The proposed approach enables the agent to exhibit a variety of tunable deceptive behaviors while ensuring the satisfaction of probabilistic constraints on the behavior. We evaluate the performance of the proposed approach via comparative user studies and present a case study on the streets of Manhattan, New York, using real travel time distributions. 
\end{quote}
\end{abstract}

\section{Introduction} \label{sec:intro}
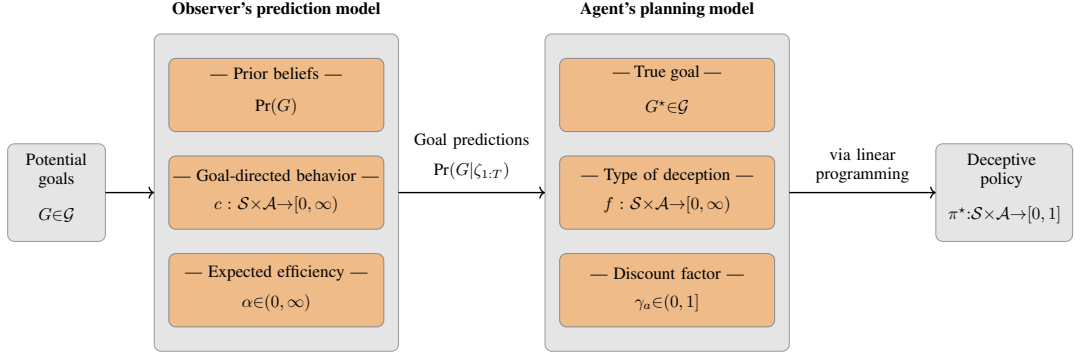
\begin{figure*}
\centering
\resizebox{0.8\textwidth}{!}{
\begin{tikzpicture}

\draw[draw=black, fill = lightgray, opacity = 0.4,rounded corners](-0.5,1.75) rectangle ++(2,2);
\node[text width = 1.5 cm,align = center] at (0.5,3.2) {Potential goals};

\node[text width = 1.5 cm,align = center] at (0.5,2.3) { $G$$\in$$\mathcal{G}$};

\draw[->, thick] (1.5,2.75) -- (2.5,2.75);

\node[] at (5,6.5) {\textbf{Observer's prediction model}};
\draw[draw=black, fill = lightgray, opacity = 0.4,rounded corners](2.5,-0.5) rectangle ++(5,6.5);
\draw[draw=black, fill = orange, opacity = 0.4,rounded corners](2.8,4) rectangle ++(4.4,1.5);
\node[text width = 3 cm,align = center] at (5,5.2) { --- Prior beliefs --- };
\node[text width = 2.8 cm,align = center] at (5,4.5) { $\text{Pr}(G)$};

\draw[draw=black, fill = orange, opacity = 0.4,rounded corners](2.8,2) rectangle ++(4.4,1.5);
\node[text width = 4.5 cm,align = center] at (5,3.1) {--- Goal-directed behavior --- };

\node[text width = 3.5 cm,align = center] at (5,2.5) {$c$ $:$ $\mathcal{S}$$\times$$\mathcal{A}$$\rightarrow$$[0,\infty)$};

\draw[draw=black, fill = orange, opacity = 0.4,rounded corners](2.8,0) rectangle ++(4.4,1.5);
\node[text width = 4 cm,align = center] at (5,1.1) {--- Expected efficiency ---};
\node[text width = 4 cm,align = center] at (5,0.5) {$\alpha$$\in$$(0,\infty)$};

\draw[->, thick] (7.5,2.75) -- (10.5,2.75);
\node[text width = 4 cm,align = center] at (9,3.8) {Goal predictions};
\node[text width = 4 cm,align = center] at (9,3.2) {$\text{Pr}(G | \zeta_{1:T})$};
\node[] at (13,6.5) {\textbf{Agent's planning model}};
\draw[draw=black, fill = lightgray, opacity = 0.4,rounded corners](10.5,-0.5) rectangle ++(5,6.5);
\draw[draw=black, fill = orange, opacity = 0.4,rounded corners](10.8,4) rectangle ++(4.4,1.5);
\node[align = center] at (13,5.2) {--- True goal ---};
\node[align = center] at (13,4.5) {$G^{\star}$$\in$$\mathcal{G}$};

\draw[draw=black, fill = orange, opacity = 0.4,rounded corners](10.8,2) rectangle ++(4.4,1.5);
\node[text width = 4 cm,align = center] at (13,3.1) {--- Type of deception --- };
\node[text width = 3 cm, align = center] at (13,2.5) {$f$ $:$ $\mathcal{S}$$\times$$\mathcal{A}$$\rightarrow$$[0,\infty)$};

\draw[draw=black, fill = orange, opacity = 0.4,rounded corners](10.8,0) rectangle ++(4.4,1.5);
\node[text width = 4 cm,align = center] at (13,1.1) {--- Discount factor ---};

\node[text width = 3 cm,align = center] at (13,0.5) {$\gamma_a$$\in$$(0,1]$};

\draw[->, thick] (15.5,2.75) -- (18.5,2.75);
\node[text width = 2 cm,align = center] at (17,3.3) {via linear programming};
\draw[draw=black, fill = lightgray, opacity = 0.4,rounded corners](18.5,1.75) rectangle ++(2.8,2);
\node[text width = 2 cm, align = center] at (19.85,3.2) {Deceptive policy};
\node[text width = 2 cm, align = center] at (19.75,2.3) { $\pi^{\star}$$:$$\mathcal{S}$$\times$$\mathcal{A}$$\rightarrow$$[0,1]$};

\end{tikzpicture}}
\caption{The overview of the proposed deceptive policy synthesis approach. Given a set $\mathcal{G}$ of potential goals, the observer's prediction model assign a probability $\text{Pr}(G|\zeta_{1:T})$ to each potential goal $G$$\in$$\mathcal{G}$ based on the agent's partial trajectory $\zeta_{1:T}$. Utilizing $\text{Pr}(G|\zeta_{1:T})$, the agent synthesizes a deceptive policy via linear programming.   }
\label{flowchart}
\end{figure*}

Deception is an important capability that is present in many human activities, ranging from sports \cite{jackson2019deception} to business \cite{chelliah2018deception} and military \cite{tsu2016art}. By making deceptive decisions, e.g., by hiding information or conveying false information, teams win games, companies secretly develop new products, and troops gain strategic advantage during battles. Although the outcomes of decisions are typically uncertain, e.g., due to incomplete knowledge and imperfect predictions, humans are still able to deceive one another effectively. 

In this paper, we develop a novel approach that enables autonomous systems to make deceptive decisions under uncertainty. Such a deception capability has the potential to improve security in adversarial environments, increase success rates in competitive settings, and create more engaging interactions in games. For example, a delivery drone may protect itself from  attacks by deceiving potentially hostile observers about its destination.


We consider an autonomous agent that carries out a task in a complex, stochastic environment. We model the agent's behavior as a Markov decision process (MDP) and express its task as reaching one of multiple potential goal states in the MDP. Being aware of the potential goals, the observer aims to predict the agent's true goal from its trajectories. The agent aims to follow a deceptive strategy that misleads the observer about the true goal  
either  by exaggerating its behavior towards a decoy goal or 
by creating ambiguity.

The main contribution of this paper is a novel approach that systematically generates globally optimal deceptive strategies in stochastic environments by combining the principle of maximum entropy with stochastic control. The proposed approach involves a number of parameters that enables the agent to exhibit tunable deceptive behaviors and allows the integration of probabilistic resource constraints into the formulation. An overview of the proposed approach is shown in Fig. \ref{flowchart}.

We express the observer's predictions on the agent's true goal by developing a prediction model based on the principle of maximum entropy \cite{ziebart2008maximum,ziebart2010modeling}. The model is based on three factors, namely, the observer's prior beliefs on the agent's true goal, a cost function expressing the agent's expected goal-directed behavior, and a constant expressing how much efficiency the observer expects from the agent. 

We synthesize deceptive strategies for the agent by developing a planning model based on stochastic optimal control \cite{puterman2014markov}. The model takes the observer's predictions as an input and constructs a constrained optimization problem that is solved via linear programming. The model is based on three factors, namely, the agent's true goal, a function expressing the type of deception, e.g., exaggeration or ambiguity, and a discount factor controlling the trade-off between the trajectory length and deception. 

We present three experiments. Firstly, we illustrate the effects of different parameters in the proposed approach on the agent's deceptive behavior. Secondly, we present online user studies and compare the proposed approach to two recently proposed deception methods \cite{masters2017deceptive,dragan2015deceptive} as well as a baseline. Finally, we present a large-scale case study on the streets of Manhattan, New York with real travel time distributions and illustrate the use of deception in realistic scenarios under probabilistic constraints on travel time. 

\subsubsection{Related Work} Deception in autonomous systems has been studied in the literature from different perspectives. In \cite{masters2017deceptive}, the authors generate deceptive plans in deterministic environments. For exaggerated behaviors, their method corresponds to a simple heuristic, i.e., reaching a decoy goal before reaching the true goal. In \cite{dragan2015deceptive}, a robot with deterministic dynamics is considered and deceptive trajectories are generated using an approach based on functional gradient descent. The work \cite{kulkarni2019unified} synthesizes obfuscated plans in deterministic environments by exploiting observation sequences. These approaches are different from the one proposed in this paper as they consider deterministic systems and the synthesized strategies are based on heuristics or local approaches. 

In \cite{ornik2018deception}, the authors synthesize deceptive strategies by expressing the evolution of observer predictions as a stochastic transition system over potential goals, which is constructed using the agent's relative distance to potential goals. Unlike  \cite{ornik2018deception}, we generate observer predictions as probability distributions over potential goals using the principle of maximum entropy. Nature-inspired deception strategies for social robots are developed in \cite{shim2012biologically,pettinati2019push}. Although these approaches are effective, their generality is limited as they lack a mathematical foundation. 

Deception has also been studied from the perspective of game theory. In \cite{wagner2011acting} and \cite{nguyen2019deception}, the authors generate deceptive strategies in single stage games and finitely repeated games, respectively. These strategies are different from the ones synthesized in this work as we focus on stochastic and dynamic settings. The works \cite{anwar2020game,cceker2016deception,kulkarni2020deceptive} study deception for cybersecurity using game-theoretic formulations. The proposed strategies are, in general, restricted to small-scale problems due to the complexity of computing equilibria in dynamic games.


\section{Background}
We model the agent's behavior in a stochastic environment as a Markov decision process (MDP). An MDP is a tuple $\mathcal{M}$$=$$(\mathcal{S},s_1,\mathcal{A},P)$ where $\mathcal{S}$ is a finite set of states, $s_1$ is a unique initial state, $\mathcal{A}$ is a finite set of actions, and $P$$:$$\mathcal{S}$$\times$$\mathcal{A}$$\times$$\mathcal{S}$$\rightarrow$$[0,1]$ is a transition function such that $\sum_{s'\in \mathcal{S}}P(s,a,s')$$=$$1$ for all $s$$\in$$\mathcal{S}$ and $a$$\in$$\mathcal{A}$. In an MDP, the agent follows a policy to achieve a task. Formally, a policy $\pi$$:$$\mathcal{S}$$\times$$\mathcal{A}$$\rightarrow$$[0,1]$ is a mapping such that $\sum_{a\in \mathcal{A}}\pi(s,a)$$=$$1$ for all $s$$\in$$\mathcal{S}$. We denote the set of all possible policies by $\Pi$.

We aim to develop an algorithm such that the agent reaches its goal in an environment while deceiving an outside observer about its goal. Hence, we consider a set of potential goals $\mathcal{G} $$\subset$$\mathcal{S}$ in the MDP and denote the agent's true goal by $G^\star$$\in$$\mathcal{G}$. 
For simplicity, we assume that all potential goal states are absorbing, i.e., $P(G,a,G)$$=$$1$ for all $G$$\in$$\mathcal{G}$.

A trajectory $\zeta$ is a sequence $(s_1,a_1,s_2,a_2,s_3,\ldots)$ of states and actions that satisfy $P(s_t,a_t,s_{t+1})$$>$$0$ for all $t$$\in$$\mathbb{N}$. A \textit{partial} trajectory $\zeta_{1:T}$ of length $T$$\in$$\mathbb{N}$ is a sequence $(s_1,a_1,s_2,\ldots,s_T)$. Let $\mathcal{T}_{\pi}$ denote the set of all admissible trajectories that are generated under the policy $\pi$, and $\zeta[t]$$:=$$s_t$ denote the state visited at the $t$-th step along $\zeta$. For a given goal state $G$$\in$$\mathcal{G}$ and a policy $\pi$, we denote by
\begin{align*}
    \text{Pr}^{\pi}(Reach[G]):=\text{Pr}\{\zeta\in\mathcal{T}_{\pi}: \exists t\in \mathbb{N}, \zeta[t]=G \}
\end{align*}
the probability with which the agent reaches the goal $G$ under the policy $\pi$. Furthermore, we denote by 
\begin{align*}
    R_{\max}(G) := \max_{\pi\in\Pi} \text{Pr}^{\pi}(Reach[G])
\end{align*}
the \textit{maximum} probability of reaching the goal $G$ under any policy. We note that the value of $R_{\max}(G)$ can be efficiently computed via value iteration \cite{Model_checking}.

For an MDP $\mathcal{M}$, let $G_{\mathcal{M}}$$=$$(\mathcal{S},E_{\mathcal{M}})$ be a directed graph where $\mathcal{S}$ is the set of vertices and $E_{\mathcal{M}}$$\subseteq$$\mathcal{S}$$\times$$\mathcal{S}$ is the set of edges such that
$(s,s')$$\in$$E_{\mathcal{M}}$ if and only if  $\sum_{a\in\mathcal{A}}P_{s,a,s'}$$>$$0$.
For the graph $G_{\mathcal{M}}$, we denote by $T_{\min}(s)$ the length of the shortest partial trajectory $\zeta_{1:T_{\min}(s)}$ such that $s_{T_{\min}(s)}$$=$$s$. Informally, $T_{\min}(s)$ indicates the minimum number of steps to reach the state $s$ from the initial state $s_1$. We use the convention $T_{\min}(s)$$=$$\infty$ if the state $s$ is not reachable from the initial state. Note that $T_{\min}(s)$ can be efficiently computed, e.g., using Dijkstra's algorithm \cite{dijkstra1959note}.

\section{Modeling Observer Predictions}

To deceive an observer about its true goal, the agent should know how the observer associates the agent's partial trajectories with potential goals. In this section, we provide a prediction model that formally expresses the observer's inference method using the principle of maximum entropy. Specifically, we present a prediction model that assigns a probability $\text{Pr}(G|\zeta_{1:T})$ to each potential goal $G$$\in$$\mathcal{G}$ for a given partial trajectory $\zeta_{1:T}$. 

An overview of the observer's prediction model is shown in Fig. \ref{flowchart}. We formally characterize the observer with its prior beliefs $\text{Pr}(G)$ on the agent's true goal, the cost function $c$$:$$\mathcal{S}$$\times$$\mathcal{A}$$\rightarrow$$[0,\infty)$ that expresses the agent's expected goal-directed behavior from the observer's perspective, and the efficiency parameter $\alpha$$\in$$(0,\infty)$ that expresses the agent's expected degree of optimality from the observer's perspective.


The prior beliefs constitute a probability distribution over the agent's potential goals and formalize \textit{where} the observer expects the agent to reach when the agent is at the initial state $s_1$. When the observer interacts with the agent only once, the prior beliefs are typically represented by a uniform distribution. In repeated interactions, Bayesian approaches can be used to construct the prior beliefs from historical data \cite{ziebart2009planning}. Note that, given the prior beliefs, we can express the observer's predictions $\text{Pr}(G|\zeta_{1:T})$ as
\begin{align}
   \text{Pr}(G | \zeta_{1:T})  =  \frac{\text{Pr}(\zeta_{1:T} | G)\text{Pr}(G)}{\sum_{G'\in\mathcal{G}}\text{Pr}(\zeta_{1:T} | G') \text{Pr}(G')},\label{bayes_eqn}
\end{align}
where
$\text{Pr}(\zeta_{1:T} | G)$ denotes the probability with which the agent follows a partial trajectory $\zeta_{1:T}$ for reaching the goal $G$. In other words, the probability $\text{Pr}(\zeta_{1:T} | G)$ expresses \textit{how} the observer expects the agent to reach a goal. We formally characterize the probability $\text{Pr}(\zeta_{1:T} | G)$ using the cost function $c$ and the efficiency parameter $\alpha$. Specifically, to reach a goal $G$$\in$$\mathcal{G}$, we assume that the observer \textit{expects} the agent to follow a policy $\overline{\pi}_G$$\in$$\Pi$ that satisfies
\begin{align*}
    \overline{\pi}_G\in \arg &\min_{\pi\in \Pi}\ \mathbb{E}^{\pi}\Bigg[\sum_{t=1}^{\infty}\gamma_o^{t-1} \Big(c(s_t,a_t)- \alpha H(\pi(s_t, \cdot))\Big)\Bigg]\\
    & \text{s.t.} \ \  \ \   \text{Pr}^{\pi}(Reach[G]) = R_{\max}(G).
\end{align*}
In the above equation, the term $H(\pi(s_t,\cdot))$ measures the entropy of the policy $\pi$ in the state $s_t$$\in$$S$ and is defined as $H(\pi(s_t,\cdot))$$=$$-\sum_{a\in\mathcal{A}}\pi(s_t,a)\log \pi(s_t,a)$. The entropy term quantifies the randomness in the agent's policy and enables the observer to reason about suboptimal trajectories. The parameter $\gamma_o$$\in$$(0,1)$ is the observer's discount factor, which is introduced only to ensure the finiteness of the solution and can be chosen arbitrarily close to one. 

The cost function $c$ expresses the expected goal-directed behavior of a rational agent. For example, in motion planning, the cost function corresponds to the distance between state pairs as observers typically expect the agent to reach its goal through the shortest feasible trajectory \cite{gergely1995taking}. We make the standard assumption \cite{dragan2013legibility,sarath21,masters21} that the cost function $c$ is known to the agent. In scenarios that involve a cooperative observer, the cost function $c$ can also be learned from demonstrations using existing learning approaches \cite{ziebart2008maximum}.

The parameter $\alpha$$\in$$(0,\infty)$ controls how much efficiency the observer expects from the agent. For example, as $\alpha$$\rightarrow$$0$, the agent is expected to be perfectly efficient and follow only the trajectories that minimize its total cost. On the other extreme, as $\alpha$$\rightarrow$$\infty$, the agent is expected to have no efficiency concerns and reach its goal by following random trajectories.
We assume that the parameter $\alpha$ is also known to the agent. In practice, one can incorporate the parameter $\alpha$ into the cost function $c$ by defining the costs as $\widetilde{c}(s,a)$$=$$c(s,a)/\alpha$ and learn the function $\widetilde{c}$ from demonstrations.

We can now derive the observer's prediction model from the agent's \textit{expected} policy $\overline{\pi}_G$ as follows. It is known \cite{haarnoja2017,ziebart2009planning} that the policy $\overline{\pi}_G$ satisfies $\overline{\pi}_G(s,a)$$=$$e^{(Q_G(s,a)-V_G(s))/\alpha}$ where 
\begin{align*}
    Q_G(s,a) &= - c(s,a) +  \gamma_o\sum_{s'\in \mathcal{S}}P(s,a,s')V_G(s')\\
    V_G(s) &= \operatornamewithlimits{softmax}\limits_{a} Q_G(s,a).
\end{align*}
In the above equations, the softmax operator is defined as $\operatorname{softmax}_x f(x)$$=$$\alpha\log \sum_x e^{f(x)/ \alpha}$. The values of $V_G(s)$ and $Q_G(s,a)$ can be iteratively computed via softmax value iteration using the initialization $V_G(G)$$=$$0$ and $V_G(s)$$=$$-C$ for all $s$$\in$$\mathcal{S}$$\backslash$$\{G\}$, where $C$ is an arbitrarily large constant. 


It is known \cite{ziebart2008maximum} that 
$\text{Pr}(\zeta_{1:T} | G)$ satisfies
\begin{align*}
    \text{Pr}(\zeta_{1:T} | G) \approx \frac{e^{-\frac{1}{\alpha}\sum_{t=1}^T c(s_t,a_t)+V_G(s_T)}}{e^{V_G(s_1)}} \prod_{t=1}^T P(s_t,a_t,s_{t+1})
\end{align*}
when the transition randomness has a limited effect on the agent's behavior and the discount factor $\gamma_o$ is large enough.
Note that for MDPs with deterministic transitions, the above expression implies that $ \text{Pr}(\zeta| G)$$\propto$$e^{-\frac{1}{\alpha}\sum_{t=1}^{\infty} c(s_t,a_t)}$. Hence, in the maximum entropy distribution, the probability of a trajectory exponentially decreases with increasing total cost. Finally, plugging $ \text{Pr}(\zeta_{1:T} | G)$ into \eqref{bayes_eqn} and simplifying terms, 
we obtain the observer's prediction model as
\begin{align}
   \text{Pr}(G | \zeta_{1:T})  \approx \frac{e^{V_G(s_T)-V_G(s_1)} \text{Pr}(G)}{\sum_{G'\in\mathcal{G}}e^{V_{G'}(s_T)-V_{G'}(s_1)} \text{Pr}(G')}.
   \label{predict_eqn}
\end{align}


Note that the observer's prediction $\text{Pr}(G | \zeta_{1:T})$ is only a function of the agent's initial state $s_1$ and the current state $s_T$, i.e., $\text{Pr}(G | \zeta_{1:T})$$=$$\text{Pr}(G |s_1,s_T)$. Hence, the observer's predictions can be computed \textit{offline} by computing the value of $V_G(s)$ for all $G$$\in$$\mathcal{G}$ and $s$$\in$$S$. This computation can be performed by running the softmax value iteration $\lvert \mathcal{G} \rvert$ times.


As the efficiency parameter $\alpha$$\rightarrow$$\infty$, for any given partial trajectory $\zeta_{1:T}$, we have $\text{Pr}(G | \zeta_{1:T})$$=$$\text{Pr}(G' | \zeta_{1:T})$. This implies that, if the observer expects the agent's goal-directed behavior to be inefficient, then the observer predicts all goals to be equally likely even after the agent's partial trajectory is revealed. In such a scenario, it is impossible to mislead the observer about the true goal. Accordingly, we will see in the experiments that the agent's deceptive behavior corresponds to reaching the true goal via shortest trajectories when the observer expects the agent to be inefficient.

\section{Synthesizing Deceptive Policies}
Being aware of the observer's prediction model, the agent aims to synthesize a policy that deceives the observer about its true goal $G^{\star}$. Formally, we propose to synthesize a deceptive policy $\pi^{\star}$$\in$$\Pi$ under which the agent maximizes the deceptiveness of its trajectory while reaching its true goal with maximum probability, i.e.,
\begin{subequations}
\begin{align}
    \pi^{\star}\in \arg &\min_{\pi\in \Pi}\ \mathbb{E}^{\pi}\Bigg[\sum_{t=1}^{\infty}g(s_t,a_t)\Bigg]\label{deceptive_obj}\\
    & \text{s.t.} \ \  \ \   \text{Pr}^{\pi}(Reach[G^{\star}]) = R_{\max}(G^{\star})\label{deceptive_cons}.
\end{align}
\end{subequations}
In \eqref{deceptive_obj}-\eqref{deceptive_cons}, we express the agent's deception objective through the generic cost function $g$$:$$\mathcal{S}$$\times$$\mathcal{A}$$\rightarrow$$[0,\infty)$. In particular, we consider a class of functions of the form 
\begin{align}
\label{generic_func_form}
    g(s,a) = \gamma_a^{T_{\min}(s)}f(s,a)
\end{align}
where $\gamma_a$$\in$$(0,1]$ is a discount factor and $f$$:$$\mathcal{S}$$\times$$\mathcal{A}$$\rightarrow$$[0,\infty)$ is a mapping that formalizes the type of deception. Recall that the constant $T_{\min}(s)$ is the minimum number of steps to reach the state $s$ from the initial state $s_1$ in the graph $G_{\mathcal{M}}$. We introduce the term $\gamma_a^{T_{\min}(s)}$ in \eqref{generic_func_form} as a scaling factor to obtain tunable agent behavior. As we will see in the experiments, as $\gamma_a$ decreases, the cost for states that are further away from the initial state becomes smaller, which encourages the agent to follow longer trajectories for deception.



\subsection{Mathematical Representation of Deception}
We design the mapping $f$ to achieve two common types of deception, namely, exaggeration and ambiguity.
\subsubsection{Exaggeration:} One of the most common strategies to deceive an observer about the true goal is exaggeration  \cite{dragan2015deceptive}. In this strategy, the agent exhibits an exaggerated behavior by pretending to reach a decoy goal, i.e., a goal that is not the true goal. 
We express the exaggeration behavior by defining $f$ as 
\begin{align}
\label{exaggeration_cost_2}
    &f(s,a)=1 + \text{Pr}(G^{\star} | s_1, s) - \max_{G \in \mathcal{G}\backslash \{G^{\star}\}} \text{Pr}(G | s_1, s)
\end{align}
if $s$$\in$$\mathcal{S}\backslash\mathcal{G}$, and $f(s,a)$$=$$0$ otherwise. 

In \eqref{exaggeration_cost_2}, the value of $f(s,a)$ linearly increases with the difference $\text{Pr}(G^{\star} | s_1, s)$$-$$\max_{G \in \mathcal{G}\backslash \{G^{\star}\}} \text{Pr}(G | s_1, s)$, i.e., the relative likelihood of the true goal with respect to a decoy goal. Hence, the smaller the value of $f(s,a)$, the more likely it is for the agent to reach a decoy goal. Additionally, we have $f(s,a)$$=$$0$ if $\text{Pr}(G^{\star} | s_1, s)$$=$$0$ and $\text{Pr}(G | s_1, s)$$=$$1$ for some $G$$\in$$\mathcal{G}\backslash \{G^{\star}\}$. That is, the agent incurs no cost in a state if the observer almost surely expects the agent to reach a decoy goal from that state. 
\subsubsection{Ambiguity:} Another possible strategy to deceive an observer about the true goal is to behave ambiguously. In this strategy, the agent exhibits an ambiguous behavior by keeping the likelihood of all potential goals similar along its trajectory. Similar to exaggeration, we express ambiguity by defining the mapping $f$ as
\begin{align}
\label{ambiguity_cost}
    f(s,a) = \sum_{G\in\mathcal{G}}\sum_{G'\in\mathcal{G}}\Big\lvert \text{Pr}(G | s_1, s)- \text{Pr}(G' | s_1, s)\Big\rvert
\end{align}
if $s$$\in$$\mathcal{S}\backslash\mathcal{G}$, and $f(s,a)$$=$$0$ otherwise.

In \eqref{ambiguity_cost}, the value of $f(s,a)$ at a state $s$ increases as the relative likelihood of a goal with respect to any other one increases. Hence, the smaller the value of $f(s,a)$, the less likely it is for the agent to try and reach a specific goal. Additionally, we have $f(s,a)$$=$$0$ if $\text{Pr}(G| s_1, s)$$=$$\text{Pr}(G'| s_1, s)$ for all $G,G'$$\in$$\mathcal{G}$, i.e., the agent incurs no cost in a state if the observer expects the agent to reach all goals equally likely. 

\subsection{Synthesis via Linear Programming}
We now synthesize deceptive policies by solving a series of linear programs (LPs). For a given MDP $\mathcal{M}$, let $\mathcal{S}_0$$\subseteq$$\mathcal{S}$ be a set of states from which there is no trajectory reaching a potential goal $G$$\in$$\mathcal{G}$. The set $\mathcal{S}_0$ can be efficiently computed through standard graph search algorithms \cite{Model_checking}. Moreover, let $\mathcal{S}_r$$=$$\mathcal{S}\backslash (\mathcal{G} \cup \mathcal{S}_0)$.
To obtain the deceptive policy $\pi^{\star}$, we first solve the following LP:
\begin{subequations}
\begin{flalign}\label{LP_start}
    &\underset{\substack{x(s,a)\geq 0}}{\text{minimize}} \sum_{s\in \mathcal{S}_r}\sum_{a\in \mathcal{A}} g(s,a) x(s,a)\\ 
    &\text{subject to:} \nonumber \\  \label{LP_cons_1}
    & \sum_{a\in \mathcal{A}}x(s,a)-\sum_{s'\in \mathcal{S}}\sum_{a\in\mathcal{A}} P(s',a,s)x(s',a)= \beta_s, \  \forall s\in \mathcal{S}_r\\  
    &\sum_{s\in \mathcal{S}_r}\sum_{a\in \mathcal{A}}x(s,a)r(s,a) = R_{\max}(G^{\star}).\label{LP_end}
\end{flalign}
\end{subequations}

In the above LP, $x(s,a)$ is a decision variable that corresponds to the agent's expected number of visits to the state-action pair $(s,a)$ \cite{puterman2014markov}. The function $\beta_s$ indicates the initial state distribution, i.e., $\beta_s$$=$$1$ if $s$$=$$s_1$, and $\beta_s$$=$$0$ otherwise. Finally, the function $r$$:$$\mathcal{S}$$\times$$\mathcal{A}$$\rightarrow$$[0,\infty)$ is the transition probability to the true goal from a given state, i.e., $r(s,a)$$=$$P(s,a,G^{\star})$ for $s$$\in$$\mathcal{S}_r$, and $r(s,a)$$=$$0$ otherwise.

The objective function in \eqref{LP_start} corresponds to the agent's expected total cost given in \eqref{deceptive_obj}. The constraint in \eqref{LP_cons_1} represents the balance equation \cite{altman1999constrained}, i.e., the expected number of times the agent enters a state is equal to the expected number of times the agent leaves that state. Finally, the constraint in \eqref{LP_end} ensures that the agent reaches its true goal $G^{\star}$ with maximum probability $R_{\max}(G^{\star})$.

It is possible to extract the deceptive policy $\pi^{\star}$ from the optimal solution of the LP in \eqref{LP_start}-\eqref{LP_end}. However, under the extracted policy, the agent may visit the states with zero cost too many times before reaching its true goal since such states do not affect the objective function. Let $v^{\star}$ be the optimal value of the LP in \eqref{LP_start}-\eqref{LP_end}. To ensure that that the agent reaches its true goal as quickly as possible while achieving its deception objective, we solve the following second LP:
\begin{subequations}
\begin{flalign}\label{LP2_start}
    &\underset{\substack{x(s,a)\geq 0}}{\text{minimize}} \ \ \ \sum_{s\in \mathcal{S}}\sum_{a\in \mathcal{A}} x(s,a)\\
    &\text{subject to:}  \ \ \sum_{s\in \mathcal{S}_r}\sum_{a\in \mathcal{A}}g(s,a)x(s,a) = v^{\star}\\
    &\qquad \qquad  \ \ \ \eqref{LP_cons_1}-\eqref{LP_end}. \label{LP2_end}
\end{flalign}
\end{subequations}
Let $\{x^{\star}(s,a)$$\geq$$0$$:$$s$$\in$$\mathcal{S}, a$$\in$$\mathcal{A}\}$ be the set of optimal variables for the LP in \eqref{LP2_start}-\eqref{LP2_end}. It follows from \cite{altman1999constrained} that the deceptive policy $\pi^{\star}$ satisfying the condition in \eqref{deceptive_obj}-\eqref{deceptive_cons} can be synthesized by choosing 
\begin{align*}\label{opt_policy_rule_approx}
   \pi^{\star}(s,a)=\begin{cases}\frac{x^{\star}(s,a)}{\sum_{a'\in\mathcal{A}}x^{\star}(s,a')} & \text{if} \ \sum_{a'\in\mathcal{A}}x^{\star}(s,a')>0,\\
   1/\lvert \mathcal{A}\rvert & \text{otherwise}.
   \end{cases}
\end{align*}

\section{Experiments}
We now demonstrate the performance of the proposed approach through numerical simulations and user studies. We run all computations on a 3.2 GHz desktop with 8 GB RAM and employ the Gurobi solver \cite{gurobi} for optimization. Necessary approvals for user studies are obtained from the appropriate institutional review boards.

\subsection{Generating Tunable Agent Behavior}
We first illustrate how to generate a range of deceptive behaviors by tuning $\alpha$ and $\gamma_a$. We consider the environment shown in Fig. \ref{tunable_figure}. The initial state is labeled with \textbf{S} and the two potential goals are labeled with \textbf{G1} and \textbf{G2}, with \textbf{G1} being the true goal. Black regions indicate the obstacles. The agent has four actions $\{right, left, up, down\}$. Under a given action, the agent transitions to the state in the corresponding direction with probability one. 

The agent's expected goal-directed behavior is to follow shortest trajectories to the goal, which we express by setting $c(s,a)$$=$$10$ for all $s$$\in$$\mathcal{S}$ and $a$$\in$$\mathcal{A}$ and $\gamma_o$$=$$0.95$. Note that any positive cost expresses the same goal-directed behavior; the value of 10 is chosen to obtain distinct behaviors for a wide range of $\alpha$ values. Recall that, as $\alpha$ gets smaller, the observer expects the agent to be more efficient and follow shorter trajectories to reach its goal. In Fig. \ref{tunable_figure} (left), the shaded region indicates all the states that a perfectly efficient agent, $\alpha$$=$$0$, can potentially visit along its trajectory to the goal \textbf{G1}. 

In Fig. \ref{tunable_figure} (left), we generate 5 trajectories to represent the agent's exaggeration behavior for various $\alpha$ and $\gamma_a$ combinations. As can be seen from the figure, for $\alpha$$\leq$$1$ and $\gamma_a$$=$$1$, the agent's exaggerated trajectory reaches the true goal while avoiding the shaded region. This trajectory is deceptive because the observer \textit{expects} the agent to be highly efficient and visit \textit{only} the states in the shaded region while reaching the goal \textbf{G1}. As $\alpha$ increases, the observer expects the agent to be less efficient. In that case, to deceive the observer, the agent starts exaggerating its behavior by getting closer to the decoy goal \textbf{G2}. As we keep increasing the $\alpha$ value, the observer expects the agent's behavior to be less goal-directed and more random. In that case, it becomes impossible to deceive the observer since any random behavior is expected. Accordingly, for $\alpha$$\geq$$20$, the agent does not try to deceive the observer and follows a shortest trajectory to its goal.

A simple heuristic to achieve exaggeration-type deceptive behavior is to first reach the decoy goal and then the true goal \cite{masters2017deceptive}. In the environment shown in Fig. \ref{tunable_figure} (left), the states that are further away from the initial state have high costs $g(s,a)$ when the discount factor is $\gamma_a$$=$$1$. Therefore, the agent has no incentive to follow longer trajectories and pretend to reach the decoy goal. However, when $\gamma_a$$=$$0.8$ and $\alpha$$=$$6$, the agent's exaggeration behavior starts exploiting those states as well and replicates the trajectory generated by the aforementioned heuristic approach.

We also generate 5 ambiguous trajectories, shown in Fig. \ref{tunable_figure} (right). In this environment, ambiguity corresponds to being at the same horizontal distance to both potential goals.
Accordingly, to achieve ambiguity for $\alpha$$\leq$$1$, the agent stays at the same horizontal distance to both potential goals for as long as possible while ensuring to visit only the states in the shaded region along its trajectory. As $\alpha$ increases, e.g., $\alpha$$=$$8$, the agent is expected to be less efficient, which enables the agent to generate ambiguity for longer. As we keep increasing the value of $\alpha$, the observer expects the agent to behave randomly. In that case, deception becomes impossible, and the agent reaches its goal by following a shortest trajectory.

\newcommand{\StaticObstacle}[2]{ \fill[] (#1+0,#2+0) rectangle  (#1+1,#2+1);}
\newcommand{\initialstate}[2]{ \fill[black!30!brown] (#1+0.1,#2+0.1) rectangle (#1+0.9,#2+0.9);}
\newcommand{\goalstate}[2]{ \fill[black!50!green] (#1+0.1,#2+0.1) rectangle (#1+0.9,#2+0.9);}
\begin{figure}[t!]
\centering
\begin{subfigure}[t]{0.22\textwidth}
\centering
\scalebox{0.36}{
\begin{tikzpicture}
\draw[black,line width=0.4pt] (0,0) grid[step=1, very thin] (9,9);
\draw[black,line width=3pt] (0,0) rectangle (9,9);

\fill[black!10!white] (4.05,0.05) rectangle (4.95,0.95);
\fill[black!10!white] (0.05,4.05) rectangle (0.95,4.95);
\fill[black!10!white] (8.05,7.05) rectangle (8.95,7.95);
\node[] at (4.5,0.5) {\Huge \textbf{S}};
\node[] at (0.5,4.5) {\Huge \textbf{G1}};
\node[] at (8.5,7.5) {\Huge \textbf{G2}};

\StaticObstacle{2}{3}
\StaticObstacle{2}{4}
\StaticObstacle{3}{3}
\StaticObstacle{3}{4}

\StaticObstacle{5}{6}
\StaticObstacle{5}{7}
\StaticObstacle{6}{6}
\StaticObstacle{6}{7}

\fill[white!80!blue] (0.05,0.05) rectangle (0.95,0.95);
\fill[white!80!blue] (1.05,0.05) rectangle (1.95,0.95);
\fill[white!80!blue] (2.05,0.05) rectangle (2.95,0.95);
\fill[white!80!blue] (3.05,0.05) rectangle (3.95,0.95);
\fill[white!80!blue] (0.05,1.05) rectangle (0.95,1.95);
\fill[white!80!blue] (1.05,1.05) rectangle (1.95,1.95);
\fill[white!80!blue] (2.05,1.05) rectangle (2.95,1.95);
\fill[white!80!blue] (3.05,1.05) rectangle (3.95,1.95);
\fill[white!80!blue] (4.05,1.05) rectangle (4.95,1.95);
\fill[white!80!blue] (0.05,2.05) rectangle (0.95,2.95);
\fill[white!80!blue] (1.05,2.05) rectangle (1.95,2.95);
\fill[white!80!blue] (2.05,2.05) rectangle (2.95,2.95);
\fill[white!80!blue] (3.05,2.05) rectangle (3.95,2.95);
\fill[white!80!blue] (4.05,2.05) rectangle (4.95,2.95);
\fill[white!80!blue] (0.05,3.05) rectangle (0.95,3.95);
\fill[white!80!blue] (1.05,3.05) rectangle (1.95,3.95);

\draw[-, line width=1mm, color = black!20!green] (4.8,0.8) -- (5.5,0.8);
\draw[-, line width=1mm, color = black!20!green] (5.5,0.8) -- (5.5,3.5);
\draw[-, line width=1mm, color = black!20!green] (5.5,3.5) -- (4.5,3.5);
\draw[-, line width=1mm, color = black!20!green] (4.5,3.5) -- (4.5,5.2);
\draw[-, line width=1mm, color = black!20!green] (4.5,5.2) -- (1.8,5.2);
\draw[-, line width=1mm, color = black!20!green] (1.8,5.2) -- (1.8,4.2);
\draw[-latex, line width=1mm, color = black!20!green] (1.8,4.2) -- (0.8,4.2);

\draw[-, line width=1mm, color = black!10!orange] (4.8,0.6) -- (6.5,0.6);
\draw[-, line width=1mm, color = black!10!orange] (6.5,0.6) -- (6.5,2.5);
\draw[-, line width=1mm, color = black!10!orange] (6.5,2.5) -- (5.8,2.5);
\draw[-, line width=1mm, color = black!10!orange] (5.8,2.5) -- (5.8,5.5);
\draw[-, line width=1mm, color = black!10!orange] (5.8,5.5) -- (1.5,5.5);
\draw[-, line width=1mm, color = black!10!orange] (1.5,5.5) -- (1.5,4.5);
\draw[-latex, line width=1mm, color = black!10!orange] (1.5,4.5) -- (0.8,4.5);

\draw[-, line width=1mm, color = white!40!blue] (4.8,0.4) -- (7.5,0.4);
\draw[-, line width=1mm, color = white!40!blue] (7.5,0.4) -- (7.5,5.8);
\draw[-, line width=1mm, color = white!40!blue] (7.5,5.8) -- (1.3,5.8);
\draw[-, line width=1mm, color = white!40!blue] (1.3,5.8) -- (1.3,4.8);
\draw[-latex, line width=1mm, color = white!40!blue] (1.3,4.8) -- (0.8,4.8);

\draw[-, line width=1mm, color = black!20!pink] (4.8,0.2) -- (7.8,0.2);
\draw[-, line width=1mm, color = black!20!pink] (7.8,0.2) -- (7.8,8.8);
\draw[-, line width=1mm, color = black!20!pink] (7.8,8.8) -- (0.5,8.8);
\draw[-latex, line width=1mm, color = black!20!pink] (0.5,8.8) -- (0.5,4.8);

\draw[-, line width=1mm, color = black!10!magenta] (4.2,0.2) -- (1.2,0.2);
\draw[-, line width=1mm, color = black!10!magenta] (1.2,0.2) -- (1.2,2.5);
\draw[-, line width=1mm, color = black!10!magenta] (1.2,2.5) -- (0.5,2.5);
\draw[-latex, line width=1mm, color = black!10!magenta] (0.5,2.5) -- (0.5,4.2);

\node[draw, fill = black!20!green!50!white] at (3.1,2.5) {\huge $\alpha$$\leq$$1$};
\draw[-latex, line width=0.5mm, color = black, opacity =0.6] (5.5,2.5) -- (4,2.5);

\node[draw, fill = black!10!orange!50!white] at (3.1,1.5) {\huge $\alpha$$=$$5$};
\draw[-latex, line width=0.5mm, color = black, opacity =0.6] (6.5,1.5) -- (4,1.5);

\node[draw, fill = white!20!blue!30!white] at (4.2,7.5) {\huge $\alpha$$=$$6$};
\draw[-latex, line width=0.5mm, color = black, opacity =0.6] (4.2,5.8) -- (4.2,7.1);

\node[draw, fill = black!20!pink!50!white, text width= 1.65 cm] at (2,7.5) {\huge $\alpha$$=$$6$ $\gamma_a$$=$$0.8$};
\draw[-latex, line width=0.5mm, color = black,  opacity =0.6] (0.5,7.5) -- (1,7.5);

\node[draw, fill = black!10!magenta!50!white] at (2.6,0.7) {\huge $\alpha$$\geq$$20$};
\draw[-latex, line width=0.5mm, color = black,  opacity =0.6] (1.2,0.7) -- (1.8,0.7);

\end{tikzpicture}}
\end{subfigure}
\begin{subfigure}[t]{0.22\textwidth}
\centering
\scalebox{0.36}{
\begin{tikzpicture}
\draw[black,line width=0.4pt] (0,0) grid[step=1, very thin] (9,9);
\draw[black,line width=3pt] (0,0) rectangle (9,9);

\fill[black!10!white] (4.05,0.05) rectangle (4.95,0.95);
\fill[black!10!white] (0.05,4.05) rectangle (0.95,4.95);
\fill[black!10!white] (8.05,7.05) rectangle (8.95,7.95);
\node[] at (4.5,0.5) {\Huge \textbf{S}};
\node[] at (0.5,4.5) {\Huge \textbf{G1}};
\node[] at (8.5,7.5) {\Huge \textbf{G2}};

\StaticObstacle{2}{3}
\StaticObstacle{2}{4}
\StaticObstacle{3}{3}
\StaticObstacle{3}{4}

\StaticObstacle{5}{6}
\StaticObstacle{5}{7}
\StaticObstacle{6}{6}
\StaticObstacle{6}{7}

\fill[white!80!blue] (0.05,0.05) rectangle (0.95,0.95);
\fill[white!80!blue] (1.05,0.05) rectangle (1.95,0.95);
\fill[white!80!blue] (2.05,0.05) rectangle (2.95,0.95);
\fill[white!80!blue] (3.05,0.05) rectangle (3.95,0.95);
\fill[white!80!blue] (0.05,1.05) rectangle (0.95,1.95);
\fill[white!80!blue] (1.05,1.05) rectangle (1.95,1.95);
\fill[white!80!blue] (2.05,1.05) rectangle (2.95,1.95);
\fill[white!80!blue] (3.05,1.05) rectangle (3.95,1.95);
\fill[white!80!blue] (4.05,1.05) rectangle (4.95,1.95);
\fill[white!80!blue] (0.05,2.05) rectangle (0.95,2.95);
\fill[white!80!blue] (1.05,2.05) rectangle (1.95,2.95);
\fill[white!80!blue] (2.05,2.05) rectangle (2.95,2.95);
\fill[white!80!blue] (3.05,2.05) rectangle (3.95,2.95);
\fill[white!80!blue] (4.05,2.05) rectangle (4.95,2.95);
\fill[white!80!blue] (0.05,3.05) rectangle (0.95,3.95);
\fill[white!80!blue] (1.05,3.05) rectangle (1.95,3.95);

\draw[-, line width=1mm, color = black!20!green] (4.2,0.8) -- (4.2,2.5);
\draw[-, line width=1mm, color = black!20!green] (4.2,2.5) -- (0.8,2.5);
\draw[-latex, line width=1mm, color = black!20!green] (0.8,2.5) -- (0.8,4.2);

\draw[-, line width=1mm, color = black!10!orange] (4.4,0.8) -- (4.4,5.5);
\draw[-, line width=1mm, color = black!10!orange] (4.4,5.5) -- (0.8,5.5);
\draw[-latex, line width=1mm, color = black!10!orange] (0.8,5.5) -- (0.8,4.8);

\draw[-, line width=1mm, color = white!40!blue] (4.6,0.8) -- (4.6,6.5);
\draw[-, line width=1mm, color = white!40!blue] (4.6,6.5) -- (0.5,6.5);
\draw[-latex, line width=1mm, color = white!40!blue] (0.5,6.5) -- (0.5,4.8);

\draw[-, line width=1mm, color = black!20!pink] (4.8,0.8) -- (4.8,7.5);
\draw[-, line width=1mm, color = black!20!pink] (4.8,7.5) -- (0.2,7.5);
\draw[-latex, line width=1mm, color = black!20!pink] (0.2,7.5) -- (0.2,4.8);

\draw[-, line width=1mm, color = black!10!magenta] (4.2,0.5) -- (0.5,0.5);
\draw[-, line width=1mm, color = black!10!magenta] (0.5,0.5) -- (0.5,2.5);
\draw[-latex, line width=1mm, color = black!10!magenta] (0.5,2.5) -- (0.5,4.2);

\node[draw, fill = black!20!green!50!white] at (2.2,1.9) {\huge $\alpha$$\leq$$1$};
\draw[-latex, line width=0.5mm, color = black, opacity = 0.6] (4.2,1.9) -- (3.2,1.9);

\node[draw, fill = black!10!orange!50!white] at (6.3,1.5) {\huge $\alpha$$=$$8$ };
\draw[-latex, line width=0.5mm, color = black, opacity = 0.6] (4.4,1.5) -- (5.5,1.5);

\node[draw, fill = white!20!blue!30!white, text width = 1.7 cm] at (6.6,2.8) {\huge $\alpha$$=$$8$, $\gamma_a$$=$$0.8$ };
\draw[-latex, line width=0.5mm, color = black, opacity = 0.6] (4.6,2.8) -- (5.6,2.8);

\node[draw, fill = black!20!pink!50!white, text width = 1.7 cm] at (6.6,4.8) {\huge $\alpha$$=$$8$, $\gamma_a$$=$$0.7$ };
\draw[-latex, line width=0.5mm, color = black, opacity = 0.6] (4.8,4.8) -- (5.5,4.8);

\node[draw, fill = black!10!magenta!50!white] at (2.2,1.1) {\huge $\alpha$$\geq$$20$};
\draw[-latex, line width=0.5mm, color = black, opacity = 0.6] (0.5,1.1) -- (1.3,1.1);

\end{tikzpicture}}
\end{subfigure}

\caption{An illustration of deceptive trajectories generated by the proposed approach under various efficiency parameters ($\alpha$) and discount factors ($\gamma_a$). The agent starts from the state {\bf{S}}. The true goal and decoy goal are {\bf{G1}} and {\bf{G2}}, respectively. $\gamma_a$$=$$1$ if it is not written explicitly. (Left) Exaggeration behavior. (Right) Ambiguity behavior.}
\label{tunable_figure}
\end{figure}
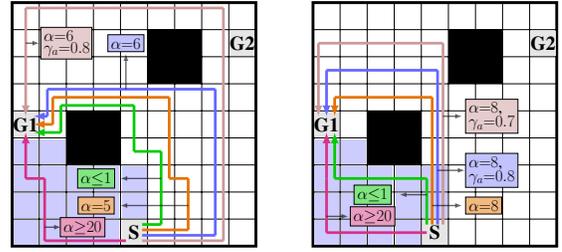

The effect of the discount factor $\gamma_a$ on ambiguity is also illustrated in Fig. \ref{tunable_figure} (right). As we decrease the value of $\gamma_a$, the cost $g(s,a)$ of the states that are further away from the initial state decreases. Consequently, the agent starts exploiting those states to achieve better ambiguity by staying at the same horizontal distance to the potential goals for longer.

\subsection{User Studies}
We conduct two user studies to evaluate the performance of the proposed approach and compare the deceptiveness of the exaggerated trajectories with a baseline and two other algorithms. We consider only exaggerated trajectories since such trajectories are known to be more deceptive than ambiguous trajectories \cite{dragan2015deceptive}. 

We consider the shortest trajectory to the true goal as the baseline (base) algorithm, which we generate by choosing $c(s,a)$$=$$1$, $\gamma_o$$=$$\gamma_a$$=$$0.95$, and $\alpha$$=$$20$ in a given environment. For comparison, we generate deceptive trajectories using the algorithms proposed in \cite{dragan2015deceptive} and \cite{masters2017deceptive}.  We note that, unlike the algorithm proposed in this paper, these algorithms are proposed for \textit{deterministic} systems and environments. 

In \cite{dragan2015deceptive}, the authors generate exaggerated (continuous) trajectories for robots. They utilize a functional gradient descent-based (GD) algorithm which \textit{locally} maximizes the cumulative goal probabilities for a decoy goal. By following \cite{dragan2015deceptive}, we initialize the GD algorithm with the baseline trajectory. In \cite{masters2017deceptive}, the authors present a deceptive path planning (DPP) algorithm to generate exaggerated trajectories by first reaching a decoy goal. In the case of multiple potential decoys, they choose the decoy goal using a heuristic which corresponds to visiting the decoy goal that is in closest distance to the true goal.


\subsubsection{Study 1: the importance of global optimality}
In the first study, we consider the $9$$\times$$9$ grid world shown in Fig. \ref{simulation_tikz2} (left). The agent starts from the state labeled with \textbf{S} and has two potential goals \textbf{G1} and \textbf{G2}, with \textbf{G1} being the true goal. Black regions indicate the obstacles.  Under each action  $a$$\in$$\{right, left, up, down\}$, the agent transitions to the state in the corresponding direction with probability one.

 Recall that the algorithm proposed in this paper, i.e., deceptive decision-making (DDM), generates \textit{globally} optimal deceptive trajectories via linear programming. In complex environments involving obstacles, as the one considered here, 
 we expect the DDM to be more deceptive than local approaches, e.g., GD. Additionally, since the decoy goal is far away from the true goal, we also expect the ``first reach a decoy goal" heuristic (as in DPP) to perform well in this environment. Hence, we hypothesize the following.
 
 $\textbf{\textup{H}}_1$: \textit{DDM and DPP generate significantly more deceptive trajectories than GD and baseline.}  
\newcommand{\Cross}{$\mathbin{\tikz [x=2ex,y=2ex,line width=.8ex, black, opacity = 1] \draw (0,0) -- (1,1) (0,1) -- (1,0);}$}%
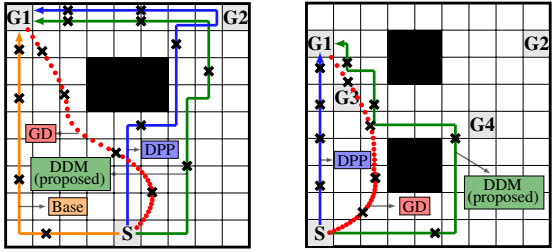
\begin{figure}[t!]
\centering
\begin{subfigure}[t!]{0.22\textwidth}
\centering
\scalebox{0.36}{
\begin{tikzpicture}
\draw[black,line width=0.4pt] (0,0) grid[step=1, very thin] (9,9);
\draw[black,line width=3pt] (0,0) rectangle (9,9);

\fill[black!10!white] (4.05,0.05) rectangle (4.95,0.95);
\fill[black!10!white] (0.05,8.05) rectangle (0.95,8.95);
\fill[black!10!white] (8.05,8.05) rectangle (8.95,8.95);

\node[] at (4.5,0.5) {\Huge \textbf{S}};
\node[] at (0.5,8.5) {\Huge \textbf{G1}};
\node[] at (8.5,8.5) {\Huge \textbf{G2}};
\StaticObstacle{3}{5}\StaticObstacle{3}{6}
\StaticObstacle{4}{5}\StaticObstacle{4}{6}
\StaticObstacle{5}{5}\StaticObstacle{5}{6}

\draw[-, line width=1mm, color = orange] (4.2,0.5) -- (0.5,0.5);
\draw[-latex, line width=1mm, color =orange] (0.5,0.5) -- (0.5,8);

\node[draw, fill = orange!50!white] at (2.3,1.5) {\huge Base};
\draw[-latex, line width=0.5mm, color = black, opacity = 0.6] (0.5,1.5) -- (1.5,1.5);

\node at (1.5,0.5) {\Cross};
\node at (0.5,2.5) {\Cross};
\node at (0.5,5.5) {\Cross};
\node at (0.5,7.3) {\Cross}; 

\draw[-, line width=1mm, color = black!50!green] (4.8,0.5) -- (6.7,0.5);
\draw[-, line width=1mm, color = black!50!green] (6.7,0.5) -- (6.7,5.5);
\draw[-, line width=1mm, color = black!50!green] (6.7,5.5) -- (7.5,5.5);
\draw[-, line width=1mm, color = black!50!green] (7.5,5.5) -- (7.5,8.35);
\draw[-latex, line width=1mm, color = black!50!green] (7.5,8.35) -- (1,8.35);

\node[draw, fill = black!50!green!50!white, text width = 2.7 cm] at (2.4,2.7) {\huge\ \ \ \ \  DDM (proposed)};
\draw[-latex, line width=0.5mm, color = black, opacity = 0.6] (6.7,2.7) -- (3.8,2.7);

\node at (6.7,3) {\Cross};
\node at (7.5,6.5) {\Cross};
\node at (5,8.35) {\Cross};
\node at (2.3,8.35) {\Cross}; 


\draw[-, line width=1mm, color = white!10!blue] (4.5,0.8) -- (4.5,4.5);
\draw[-, line width=1mm, color = white!10!blue] (4.5,4.5) -- (6.3,4.5);
\draw[-, line width=1mm, color = white!10!blue] (6.3,4.5) -- (6.3,8.2);
\draw[-, line width=1mm, color = white!10!blue] (6.3,8.2) -- (7.8,8.2);
\draw[-, line width=1mm, color = white!10!blue] (7.8,8.2) -- (7.8,8.75);
\draw[-latex, line width=1mm, color = white!10!blue] (7.8,8.75) -- (1,8.75);

\node[draw, fill = white!10!blue!50!white] at (5.7,3.6) {\huge DPP};
\draw[-latex, line width=0.5mm, color = black, opacity = 0.6] (4.5,3.6) -- (5,3.6);

\node at (5,4.5) {\Cross};
\node at (6.3,7.5) {\Cross};
\node at (5,8.75) {\Cross};
\node at (2.3,8.75) {\Cross};

\filldraw[color = red] (4.845,0.721) circle (2pt);
\filldraw[color = red] (4.977,0.867) circle (2pt);
\filldraw[color = red] (5.093,1.012) circle (2pt);
\filldraw[color = red] (5.191,1.157) circle (2pt);
\filldraw[color = red] (5.273,1.303) circle (2pt);
\filldraw[color = red] (5.338,1.448) circle (2pt);
\filldraw[color = red] (5.384,1.594) circle (2pt);
\filldraw[color = red] (5.412,1.739) circle (2pt);
\filldraw[color = red] (5.420,1.884) circle (2pt);
\filldraw[color = red] (5.410,2.030) circle (2pt);
\filldraw[color = red] (5.378,2.175) circle (2pt);
\filldraw[color = red] (5.327,2.321) circle (2pt);
\filldraw[color = red] (5.253,2.466) circle (2pt);
\filldraw[color = red] (5.158,2.612) circle (2pt);
\filldraw[color = red] (5.039,2.757) circle (2pt);
\filldraw[color = red] (4.897,2.902) circle (2pt);
\filldraw[color = red] (4.730,3.048) circle (2pt);
\filldraw[color = red] (4.538,3.193) circle (2pt);
\filldraw[color = red] (4.320,3.339) circle (2pt);
\filldraw[color = red] (4.075,3.484) circle (2pt);
\filldraw[color = red] (3.803,3.630) circle (2pt);
\filldraw[color = red] (3.511,3.781) circle (2pt);
\filldraw[color = red] (3.215,3.940) circle (2pt);
\filldraw[color = red] (2.916,4.113) circle (2pt);
\filldraw[color = red] (2.656,4.325) circle (2pt);
\filldraw[color = red] (2.442,4.574) circle (2pt);
\filldraw[color = red] (2.296,4.840) circle (2pt);
\filldraw[color = red] (2.278,5.107) circle (2pt);
\filldraw[color = red] (2.237,5.373) circle (2pt);
\filldraw[color = red] (2.171,5.640) circle (2pt);
\filldraw[color = red] (2.080,5.906) circle (2pt);
\filldraw[color = red] (1.967,6.172) circle (2pt);
\filldraw[color = red] (1.833,6.439) circle (2pt);
\filldraw[color = red] (1.683,6.705) circle (2pt);
\filldraw[color = red] (1.520,6.971) circle (2pt);
\filldraw[color = red] (1.350,7.238) circle (2pt);
\filldraw[color = red] (1.174,7.504) circle (2pt);
\filldraw[color = red] (0.997,7.770) circle (2pt);
\filldraw[color = red] (0.820,8.037) circle (2pt);
\node at (5.410,2.030) {\Cross};
\node at (4.075,3.484) {\Cross};
\node at (2.171,5.640) {\Cross};
\node at (1.350,7.238) {\Cross};

\node[draw, fill = red!50!white] at (1.3,4.2) {\huge GD};
\draw[-latex, line width=0.5mm, color = black, opacity = 0.6] (2.7,4.2) -- (1.8,4.2);


\end{tikzpicture}}
\end{subfigure}
\begin{subfigure}[t!]{0.22\textwidth}
\centering
\scalebox{0.36}{
\begin{tikzpicture}
\draw[black,line width=0.4pt] (0,0) grid[step=1, very thin] (9,9);
\draw[black,line width=3pt] (0,0) rectangle (9,9);

\fill[black!10!white] (0.05,0.05) rectangle (0.95,0.95);
\fill[black!10!white] (0.05,7.05) rectangle (0.95,7.95);
\fill[black!10!white] (8.05,7.05) rectangle (8.95,7.95);
\fill[black!10!white] (1.05,5.05) rectangle (1.95,5.95);
\fill[black!10!white] (6.05,4.05) rectangle (6.95,4.95);
\node[] at (0.5,0.5) {\Huge \textbf{S}};
\node[] at (0.5,7.5) {\Huge \textbf{G1}};
\node[] at (8.5,7.5) {\Huge \textbf{G2}};
\node[] at (1.5,5.5) {\Huge \textbf{G3}};
\node[] at (6.5,4.5) {\Huge \textbf{G4}};
\StaticObstacle{3}{6}\StaticObstacle{3}{7}
\StaticObstacle{4}{6}\StaticObstacle{4}{7}

\StaticObstacle{3}{2}\StaticObstacle{3}{3}
\StaticObstacle{4}{2}\StaticObstacle{4}{3}

\draw[-, line width=1mm, color = black!50!green] (0.8,0.5) -- (5.5,0.5);
\draw[-, line width=1mm, color = black!50!green] (5.5,0.5) -- (5.5,4.5);
\draw[-, line width=1mm, color = black!50!green] (5.5,4.5) -- (2.5,4.5);
\draw[-, line width=1mm, color = black!50!green] (2.5,4.5) -- (2.5,6.5);
\draw[-, line width=1mm, color = black!50!green] (2.5,6.5) -- (1.5,6.5);
\draw[-, line width=1mm, color = black!50!green] (1.5,6.5) -- (1.5,7.5);
\draw[-latex, line width=1mm, color = black!50!green] (1.5,7.5) -- (1,7.5);

\node[draw, fill = black!50!green!50!white, text width = 2.7 cm] at (7.3,2) {\huge\ \ \ \ \  DDM (proposed)};
\draw[-latex, line width=0.5mm, color = black, opacity =0.6] (5.5,3.5) -- (6.8,2.7);

\node at (4.75,0.5) {\Cross}; 
\node at (5.5,4) {\Cross};
\node at (2.5,5.25) {\Cross}; 
\node at (1.5,6.8) {\Cross}; 

\draw[-latex, line width=1mm, color = white!10!blue] (0.5,0.8) -- (0.5,7.2);

\node[draw, fill = white!10!blue!50!white] at (1.7,3.2) {\huge DPP};
\draw[-latex, line width=0.5mm, color = black, opacity = 0.6] (0.5,3.2) -- (1,3.2);

\node at (0.5,2.25) {\Cross};
\node at (0.5,4) {\Cross};
\node at (0.5,5.25) {\Cross};
\node at (0.5,6.6) {\Cross};

\filldraw[color = red] (0.854,0.518) circle (2pt);
\filldraw[color = red] (1.020,0.586) circle (2pt);
\filldraw[color = red] (1.179,0.657) circle (2pt);
\filldraw[color = red] (1.331,0.733) circle (2pt);
\filldraw[color = red] (1.475,0.812) circle (2pt);
\filldraw[color = red] (1.612,0.895) circle (2pt);
\filldraw[color = red] (1.740,0.982) circle (2pt);
\filldraw[color = red] (1.861,1.073) circle (2pt);
\filldraw[color = red] (1.972,1.169) circle (2pt);
\filldraw[color = red] (2.075,1.268) circle (2pt);
\filldraw[color = red] (2.169,1.373) circle (2pt);
\filldraw[color = red] (2.254,1.481) circle (2pt);
\filldraw[color = red] (2.329,1.595) circle (2pt);
\filldraw[color = red] (2.394,1.714) circle (2pt);
\filldraw[color = red] (2.449,1.837) circle (2pt);
\filldraw[color = red] (2.492,1.966) circle (2pt);
\filldraw[color = red] (2.522,2.100) circle (2pt);
\filldraw[color = red] (2.541,2.240) circle (2pt);
\filldraw[color = red] (2.548,2.385) circle (2pt);
\filldraw[color = red] (2.543,2.537) circle (2pt);
\filldraw[color = red] (2.524,2.696) circle (2pt);
\filldraw[color = red] (2.490,2.861) circle (2pt);
\filldraw[color = red] (2.509,3.033) circle (2pt);
\filldraw[color = red] (2.512,3.213) circle (2pt);
\filldraw[color = red] (2.497,3.400) circle (2pt);
\filldraw[color = red] (2.489,3.596) circle (2pt);
\filldraw[color = red] (2.486,3.800) circle (2pt);
\filldraw[color = red] (2.463,4.013) circle (2pt);
\filldraw[color = red] (2.419,4.236) circle (2pt);
\filldraw[color = red] (2.355,4.470) circle (2pt);
\filldraw[color = red] (2.270,4.713) circle (2pt);
\filldraw[color = red] (2.163,4.967) circle (2pt);
\filldraw[color = red] (2.035,5.233) circle (2pt);
\filldraw[color = red] (1.886,5.508) circle (2pt);
\filldraw[color = red] (1.716,5.794) circle (2pt);
\filldraw[color = red] (1.530,6.088) circle (2pt);
\filldraw[color = red] (1.332,6.388) circle (2pt);
\filldraw[color = red] (1.126,6.692) circle (2pt);
\filldraw[color = red] (0.918,6.997) circle (2pt);
\node at (2.075,1.268) {\Cross};
\node at (2.543,2.537) {\Cross};
\node at (2.355,4.470) {\Cross};
\node at (1.530,6.088) {\Cross};

\node[draw, fill = white!50!red] at (4,1.5) {\huge GD};
\draw[-latex, line width=0.5mm, color = black, opacity = 0.6] (2.3,1.5) -- (3.3,1.5);

\end{tikzpicture}}
\end{subfigure}

\caption{Environments and trajectories in the user studies. Crosses indicate the points up to which a trajectory is shown to the users. In study 2, the baseline is the same with DPP. (Left) User study 1. (Right) User study 2.  }
\label{simulation_tikz2}
\end{figure}

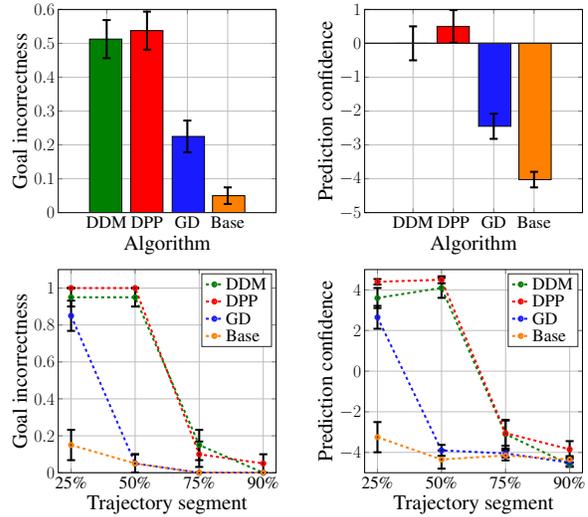
\begin{figure}[t!]
\vspace{-2cm}
\centering
\begin{subfigure}[t!]{0.5\textwidth}
\centering
\scalebox{0.38}{
%
%
\definecolor{mycolor1}{rgb}{1.00000,0.00000,1.00000}%
\begin{tikzpicture}

\begin{axis}[%
width=3in,
height=2.8in,
at={(0.28in,0.662in)},
scale only axis,
bar shift auto,
xmin=-0.2,
xmax=5.2,
xtick={1,2,3,4},
xticklabels={{DDM},{DPP},{GD},{Base}},
xlabel style={font=\color{white!15!black},font=\Huge},
xticklabel style = {font=\huge},
xlabel={Algorithm},
ymin=0,
ymax=0.6,
ylabel style={font=\color{white!15!black},font=\Huge, align = center},
ylabel={Goal incorrectness},
yticklabel style = {font=\huge},
axis background/.style={fill=white},
xmajorgrids,
ymajorgrids
]
\addplot[ybar, bar width=0.8, fill=black!50!green, draw=black, area legend] table[row sep=crcr] {%
2.25	0.5125\\
};
\addplot[forget plot, color=white!15!black] table[row sep=crcr] {%
-0.2	0\\
5.2	0\\
};
\addplot [color=black, line width=2.0pt, forget plot]
 plot [error bars/.cd, y dir=both, y explicit, error bar style={line width=2.0pt}, error mark options={line width=2.0pt, mark size=4.0pt, rotate=90}]
 table[row sep=crcr, y error plus index=2, y error minus index=3]{%
1	0.5125	0.0562368128001939	0.0562368128001939\\
};
\addplot[ybar, bar width=0.8, fill=red, draw=black, area legend] table[row sep=crcr] {%
2.42	0.5375\\
};
\addplot[forget plot, color=white!15!black] table[row sep=crcr] {%
-0.2	0\\
5.2	0\\
};
\addplot [color=black, line width=2.0pt, forget plot]
 plot [error bars/.cd, y dir=both, y explicit, error bar style={line width=2.0pt}, error mark options={line width=2.0pt, mark size=4.0pt, rotate=90}]
 table[row sep=crcr, y error plus index=2, y error minus index=3]{%
2	0.5375	0.056095956441743	0.056095956441743\\
};
\addplot[ybar, bar width=0.8, fill=white!10!blue, draw=black, area legend] table[row sep=crcr] {%
2.6	0.225\\
};
\addplot[forget plot, color=white!15!black] table[row sep=crcr] {%
-0.2	0\\
5.2	0\\
};
\addplot [color=black, line width=2.0pt, forget plot]
 plot [error bars/.cd, y dir=both, y explicit, error bar style={line width=2.0pt}, error mark options={line width=2.0pt, mark size=4.0pt, rotate=90}]
 table[row sep=crcr, y error plus index=2, y error minus index=3]{%
3	0.225	0.0469816823987036	0.0469816823987036\\
};
\addplot[ybar, bar width=0.8, fill=orange, draw=black, area legend] table[row sep=crcr] {%
2.75	0.05\\
};
\addplot[forget plot, color=white!15!black] table[row sep=crcr] {%
-0.2	0\\
5.2	0\\
};
\addplot [color=black, line width=2.0pt, forget plot]
 plot [error bars/.cd, y dir=both, y explicit, error bar style={line width=2.0pt}, error mark options={line width=2.0pt, mark size=4.0pt, rotate=90}]
 table[row sep=crcr, y error plus index=2, y error minus index=3]{%
4	0.05	0.0245207223136842	0.0245207223136842\\
};
\end{axis}

\begin{axis}[%
width=3in,
height=2.8in,
at={(4.5in,0.662in)},
scale only axis,
bar shift auto,
xmin=-0.2,
xmax=5.2,
xtick={1,2,3,4},
xticklabels={{DDM},{DPP},{GD},{Base}},
xlabel style={font=\color{white!15!black},font=\Huge},
xticklabel style = {font=\huge},
xlabel={Algorithm},
ymin=-5,
ymax=1,
ylabel style={font=\color{white!15!black},font=\Huge},
ylabel={Prediction confidence},
yticklabel style = {font=\huge},
axis background/.style={fill=white},
xmajorgrids,
ymajorgrids
]
\addplot[ybar, bar width=0.8, fill=black!50!green, draw=black, area legend] table[row sep=crcr] {%
2.25	0\\
};
\addplot[forget plot, color=white!15!black] table[row sep=crcr] {%
-0.2	0\\
5.2	0\\
};
\addplot [color=black, line width=2.0pt, forget plot]
 plot [error bars/.cd, y dir=both, y explicit, error bar style={line width=2.0pt}, error mark options={line width=2.0pt, mark size=4.0pt, rotate=90}]
 table[row sep=crcr, y error plus index=2, y error minus index=3]{%
1	0	0.50126422452117	0.50126422452117\\
};
\addplot[ybar, bar width=0.8, fill=red, draw=black, area legend] table[row sep=crcr] {%
2.42	0.5\\
};
\addplot[forget plot, color=white!15!black] table[row sep=crcr] {%
-0.2	0\\
5.2	0\\
};
\addplot [color=black, line width=2.0pt, forget plot]
 plot [error bars/.cd, y dir=both, y explicit, error bar style={line width=2.0pt}, error mark options={line width=2.0pt, mark size=4.0pt, rotate=90}]
 table[row sep=crcr, y error plus index=2, y error minus index=3]{%
2	0.5	0.483918603888994	0.483918603888994\\
};
\addplot[ybar, bar width=0.8, fill=white!10!blue, draw=black, area legend] table[row sep=crcr] {%
2.6	-2.45\\
};
\addplot[forget plot, color=white!15!black] table[row sep=crcr] {%
-0.2	0\\
5.2	0\\
};
\addplot [color=black, line width=2.0pt, forget plot]
 plot [error bars/.cd, y dir=both, y explicit, error bar style={line width=2.0pt}, error mark options={line width=2.0pt, mark size=4.0pt, rotate=90}]
 table[row sep=crcr, y error plus index=2, y error minus index=3]{%
3	-2.45	0.371406836198102	0.371406836198102\\
};
\addplot[ybar, bar width=0.8, fill=orange, draw=black, area legend] table[row sep=crcr] {%
2.75	-4.025\\
};
\addplot[forget plot, color=white!15!black] table[row sep=crcr] {%
-0.2	0\\
5.2	0\\
};
\addplot [color=black, line width=2.0pt, forget plot]
 plot [error bars/.cd, y dir=both, y explicit, error bar style={line width=2.0pt}, error mark options={line width=2.0pt, mark size=4.0pt, rotate=90}]
 table[row sep=crcr, y error plus index=2, y error minus index=3]{%
4	-4.025	0.230557250091209	0.230557250091209\\
};
\end{axis}

\begin{axis}[%
width=12.521in,
height=5.621in,
at={(0in,0in)},
scale only axis,
xmin=0,
xmax=1,
ymin=0,
ymax=1,
axis line style={draw=none},
ticks=none,
axis x line*=bottom,
axis y line*=left
]
\end{axis}
\end{tikzpicture}%

}
\end{subfigure}\\[-2cm]
\begin{subfigure}[t!]{0.5\textwidth}
\centering
\scalebox{0.38}{

%
%
\definecolor{mycolor1}{rgb}{1.00000,0.00000,1.00000}%
\begin{tikzpicture}

\begin{axis}[%
width=3in,
height=2.8in,
at={(0.28in,0.662in)},
scale only axis,
xmin=0.8,
xmax=4.2,
xtick={1,2,3,4},
xticklabels={{25\%},{50\%},{75\%},{90\%}},
xlabel style={font=\color{white!15!black},font=\Huge},
xticklabel style = {font=\huge},
xlabel={Trajectory segment},
ymin=0,
ymax=1.1,
ylabel style={font=\color{white!15!black},font=\Huge},
ylabel={Goal incorrectness},
yticklabel style = {font=\huge},
axis background/.style={fill=white},
xmajorgrids,
ymajorgrids,
legend style={legend cell align=left, align=left, draw=white!15!black, font=\huge}
]
\addplot [color=black!50!green, dashed, line width=2.0pt, mark=o, mark options={solid, black!50!green}]
  table[row sep=crcr]{%
1	0.95\\
2	0.95\\
3	0.15\\
4	0\\
};
\addlegendentry{DDM}

\addplot [color=red, dashed, line width=2.0pt, mark=o, mark options={solid, red}]
  table[row sep=crcr]{%
1	1\\
2	1\\
3	0.1\\
4	0.05\\
};
\addlegendentry{DPP}

\addplot [color=white!10!blue, dashed, line width=2.0pt, mark=o, mark options={solid, white!10!blue}]
  table[row sep=crcr]{%
1	0.85\\
2	0.05\\
3	0\\
4	0\\
};
\addlegendentry{GD}

\addplot [color=orange, dashed, line width=2.0pt, mark=o, mark options={solid, orange}]
  table[row sep=crcr]{%
1	0.15\\
2	0.05\\
3	0\\
4	0\\
};
\addlegendentry{Base}

\addplot [color=black, line width=2.0pt, forget plot]
 plot [error bars/.cd, y dir=both, y explicit, error bar style={line width=2.0pt}, error mark options={line width=2.0pt, mark size=4.0pt, rotate=90}]
 table[row sep=crcr, y error plus index=2, y error minus index=3]{%
1	0.95	0.05	0.05\\
};
\addplot [color=black, line width=2.0pt, forget plot]
 plot [error bars/.cd, y dir=both, y explicit, error bar style={line width=2.0pt}, error mark options={line width=2.0pt, mark size=4.0pt, rotate=90}]
 table[row sep=crcr, y error plus index=2, y error minus index=3]{%
1	1	0	0\\
};
\addplot [color=black, line width=2.0pt, forget plot]
 plot [error bars/.cd, y dir=both, y explicit, error bar style={line width=2.0pt}, error mark options={line width=2.0pt, mark size=4.0pt, rotate=90}]
 table[row sep=crcr, y error plus index=2, y error minus index=3]{%
1	0.85	0.0819178021909125	0.0819178021909125\\
};
\addplot [color=black, line width=2.0pt, forget plot]
 plot [error bars/.cd, y dir=both, y explicit, error bar style={line width=2.0pt}, error mark options={line width=2.0pt, mark size=4.0pt, rotate=90}]
 table[row sep=crcr, y error plus index=2, y error minus index=3]{%
1	0.15	0.0819178021909125	0.0819178021909125\\
};
\addplot [color=black, line width=2.0pt, forget plot]
 plot [error bars/.cd, y dir=both, y explicit, error bar style={line width=2.0pt}, error mark options={line width=2.0pt, mark size=4.0pt, rotate=90}]
 table[row sep=crcr, y error plus index=2, y error minus index=3]{%
2	0.95	0.05	0.05\\
};
\addplot [color=black, line width=2.0pt, forget plot]
 plot [error bars/.cd, y dir=both, y explicit, error bar style={line width=2.0pt}, error mark options={line width=2.0pt, mark size=4.0pt, rotate=90}]
 table[row sep=crcr, y error plus index=2, y error minus index=3]{%
2	1	0	0\\
};
\addplot [color=black, line width=2.0pt, forget plot]
 plot [error bars/.cd, y dir=both, y explicit, error bar style={line width=2.0pt}, error mark options={line width=2.0pt, mark size=4.0pt, rotate=90}]
 table[row sep=crcr, y error plus index=2, y error minus index=3]{%
2	0.05	0.05	0.05\\
};
\addplot [color=black, line width=2.0pt, forget plot]
 plot [error bars/.cd, y dir=both, y explicit, error bar style={line width=2.0pt}, error mark options={line width=2.0pt, mark size=4.0pt, rotate=90}]
 table[row sep=crcr, y error plus index=2, y error minus index=3]{%
2	0.05	0.05	0.05\\
};
\addplot [color=black, line width=2.0pt, forget plot]
 plot [error bars/.cd, y dir=both, y explicit, error bar style={line width=2.0pt}, error mark options={line width=2.0pt, mark size=4.0pt, rotate=90}]
 table[row sep=crcr, y error plus index=2, y error minus index=3]{%
3	0.15	0.0819178021909125	0.0819178021909125\\
};
\addplot [color=black, line width=2.0pt, forget plot]
 plot [error bars/.cd, y dir=both, y explicit, error bar style={line width=2.0pt}, error mark options={line width=2.0pt, mark size=4.0pt, rotate=90}]
 table[row sep=crcr, y error plus index=2, y error minus index=3]{%
3	0.1	0.0688247201611685	0.0688247201611685\\
};
\addplot [color=black, line width=2.0pt, forget plot]
 plot [error bars/.cd, y dir=both, y explicit, error bar style={line width=2.0pt}, error mark options={line width=2.0pt, mark size=4.0pt, rotate=90}]
 table[row sep=crcr, y error plus index=2, y error minus index=3]{%
3	0	0	0\\
};
\addplot [color=black, line width=2.0pt, forget plot]
 plot [error bars/.cd, y dir=both, y explicit, error bar style={line width=2.0pt}, error mark options={line width=2.0pt, mark size=4.0pt, rotate=90}]
 table[row sep=crcr, y error plus index=2, y error minus index=3]{%
3	0	0	0\\
};
\addplot [color=black, line width=2.0pt, forget plot]
 plot [error bars/.cd, y dir=both, y explicit, error bar style={line width=2.0pt}, error mark options={line width=2.0pt, mark size=4.0pt, rotate=90}]
 table[row sep=crcr, y error plus index=2, y error minus index=3]{%
4	0	0	0\\
};
\addplot [color=black, line width=2.0pt, forget plot]
 plot [error bars/.cd, y dir=both, y explicit, error bar style={line width=2.0pt}, error mark options={line width=2.0pt, mark size=4.0pt, rotate=90}]
 table[row sep=crcr, y error plus index=2, y error minus index=3]{%
4	0.05	0.05	0.05\\
};
\addplot [color=black, line width=2.0pt, forget plot]
 plot [error bars/.cd, y dir=both, y explicit, error bar style={line width=2.0pt}, error mark options={line width=2.0pt, mark size=4.0pt, rotate=90}]
 table[row sep=crcr, y error plus index=2, y error minus index=3]{%
4	0	0	0\\
};
\addplot [color=black, line width=2.0pt, forget plot]
 plot [error bars/.cd, y dir=both, y explicit, error bar style={line width=2.0pt}, error mark options={line width=2.0pt, mark size=4.0pt, rotate=90}]
 table[row sep=crcr, y error plus index=2, y error minus index=3]{%
4	0	0	0\\
};
\end{axis}

\begin{axis}[%
width=3in,
height=2.8in,
at={(4.5in,0.662in)},
scale only axis,
xmin=0.8,
xmax=4.2,
xtick={1,2,3,4},
xticklabels={{25\%},{50\%},{75\%},{90\%}},
xlabel style={font=\color{white!15!black},font=\Huge},
xlabel={Trajectory segment},
xticklabel style = {font=\huge},
ymin=-5,
ymax=5,
ylabel style={font=\color{white!15!black},font=\Huge},
ylabel={Prediction confidence},
yticklabel style = {font=\huge},
axis background/.style={fill=white},
xmajorgrids,
ymajorgrids,
legend style={legend cell align=left, align=left, draw=white!15!black, font=\huge }
]
\addplot [color=black!50!green, dashed, line width=2.0pt, mark=o, mark options={solid, black!50!green}]
  table[row sep=crcr]{%
1	3.6\\
2	4.1\\
3	-3.1\\
4	-4.6\\
};
\addlegendentry{DDM}

\addplot [color=red, dashed, line width=2.0pt, mark=o, mark options={solid, red}]
  table[row sep=crcr]{%
1	4.4\\
2	4.5\\
3	-3.05\\
4	-3.85\\
};
\addlegendentry{DPP}

\addplot [color=white!10!blue, dashed, line width=2.0pt, mark=o, mark options={solid, white!10!blue}]
  table[row sep=crcr]{%
1	2.65\\
2	-3.9\\
3	-4.05\\
4	-4.5\\
};
\addlegendentry{GD}

\addplot [color=orange, dashed, line width=2.0pt, mark=o, mark options={solid, orange}]
  table[row sep=crcr]{%
1	-3.25\\
2	-4.35\\
3	-4.15\\
4	-4.35\\
};
\addlegendentry{Base}

\addplot [color=black, line width=2.0pt, forget plot]
 plot [error bars/.cd, y dir=both, y explicit, error bar style={line width=2.0pt}, error mark options={line width=2.0pt, mark size=4.0pt, rotate=90}]
 table[row sep=crcr, y error plus index=2, y error minus index=3]{%
1	3.6	0.494176614497074	0.494176614497074\\
};
\addplot [color=black, line width=2.0pt, forget plot]
 plot [error bars/.cd, y dir=both, y explicit, error bar style={line width=2.0pt}, error mark options={line width=2.0pt, mark size=4.0pt, rotate=90}]
 table[row sep=crcr, y error plus index=2, y error minus index=3]{%
1	4.4	0.133771210811988	0.133771210811988\\
};
\addplot [color=black, line width=2.0pt, forget plot]
 plot [error bars/.cd, y dir=both, y explicit, error bar style={line width=2.0pt}, error mark options={line width=2.0pt, mark size=4.0pt, rotate=90}]
 table[row sep=crcr, y error plus index=2, y error minus index=3]{%
1	2.65	0.563237820201434	0.563237820201434\\
};
\addplot [color=black, line width=2.0pt, forget plot]
 plot [error bars/.cd, y dir=both, y explicit, error bar style={line width=2.0pt}, error mark options={line width=2.0pt, mark size=4.0pt, rotate=90}]
 table[row sep=crcr, y error plus index=2, y error minus index=3]{%
1	-3.25	0.742949314698201	0.742949314698201\\
};
\addplot [color=black, line width=2.0pt, forget plot]
 plot [error bars/.cd, y dir=both, y explicit, error bar style={line width=2.0pt}, error mark options={line width=2.0pt, mark size=4.0pt, rotate=90}]
 table[row sep=crcr, y error plus index=2, y error minus index=3]{%
2	4.1	0.491506813145475	0.491506813145475\\
};
\addplot [color=black, line width=2.0pt, forget plot]
 plot [error bars/.cd, y dir=both, y explicit, error bar style={line width=2.0pt}, error mark options={line width=2.0pt, mark size=4.0pt, rotate=90}]
 table[row sep=crcr, y error plus index=2, y error minus index=3]{%
2	4.5	0.17013926184468	0.17013926184468\\
};
\addplot [color=black, line width=2.0pt, forget plot]
 plot [error bars/.cd, y dir=both, y explicit, error bar style={line width=2.0pt}, error mark options={line width=2.0pt, mark size=4.0pt, rotate=90}]
 table[row sep=crcr, y error plus index=2, y error minus index=3]{%
2	-3.9	0.27047716121115	0.27047716121115\\
};
\addplot [color=black, line width=2.0pt, forget plot]
 plot [error bars/.cd, y dir=both, y explicit, error bar style={line width=2.0pt}, error mark options={line width=2.0pt, mark size=4.0pt, rotate=90}]
 table[row sep=crcr, y error plus index=2, y error minus index=3]{%
2	-4.35	0.448828885457535	0.448828885457535\\
};
\addplot [color=black, line width=2.0pt, forget plot]
 plot [error bars/.cd, y dir=both, y explicit, error bar style={line width=2.0pt}, error mark options={line width=2.0pt, mark size=4.0pt, rotate=90}]
 table[row sep=crcr, y error plus index=2, y error minus index=3]{%
3	-3.1	0.70300108557227	0.70300108557227\\
};
\addplot [color=black, line width=2.0pt, forget plot]
 plot [error bars/.cd, y dir=both, y explicit, error bar style={line width=2.0pt}, error mark options={line width=2.0pt, mark size=4.0pt, rotate=90}]
 table[row sep=crcr, y error plus index=2, y error minus index=3]{%
3	-3.05	0.617614255357705	0.617614255357705\\
};
\addplot [color=black, line width=2.0pt, forget plot]
 plot [error bars/.cd, y dir=both, y explicit, error bar style={line width=2.0pt}, error mark options={line width=2.0pt, mark size=4.0pt, rotate=90}]
 table[row sep=crcr, y error plus index=2, y error minus index=3]{%
3	-4.05	0.198348444065382	0.198348444065382\\
};
\addplot [color=black, line width=2.0pt, forget plot]
 plot [error bars/.cd, y dir=both, y explicit, error bar style={line width=2.0pt}, error mark options={line width=2.0pt, mark size=4.0pt, rotate=90}]
 table[row sep=crcr, y error plus index=2, y error minus index=3]{%
3	-4.15	0.243602350693005	0.243602350693005\\
};
\addplot [color=black, line width=2.0pt, forget plot]
 plot [error bars/.cd, y dir=both, y explicit, error bar style={line width=2.0pt}, error mark options={line width=2.0pt, mark size=4.0pt, rotate=90}]
 table[row sep=crcr, y error plus index=2, y error minus index=3]{%
4	-4.6	0.112390297389803	0.112390297389803\\
};
\addplot [color=black, line width=2.0pt, forget plot]
 plot [error bars/.cd, y dir=both, y explicit, error bar style={line width=2.0pt}, error mark options={line width=2.0pt, mark size=4.0pt, rotate=90}]
 table[row sep=crcr, y error plus index=2, y error minus index=3]{%
4	-3.85	0.405715741816724	0.405715741816724\\
};
\addplot [color=black, line width=2.0pt, forget plot]
 plot [error bars/.cd, y dir=both, y explicit, error bar style={line width=2.0pt}, error mark options={line width=2.0pt, mark size=4.0pt, rotate=90}]
 table[row sep=crcr, y error plus index=2, y error minus index=3]{%
4	-4.5	0.17013926184468	0.17013926184468\\
};
\addplot [color=black, line width=2.0pt, forget plot]
 plot [error bars/.cd, y dir=both, y explicit, error bar style={line width=2.0pt}, error mark options={line width=2.0pt, mark size=4.0pt, rotate=90}]
 table[row sep=crcr, y error plus index=2, y error minus index=3]{%
4	-4.35	0.166622801245018	0.166622801245018\\
};
\end{axis}

\begin{axis}[%
width=12.49in,
height=5.621in,
at={(0in,0in)},
scale only axis,
xmin=0,
xmax=1,
ymin=0,
ymax=1,
axis line style={draw=none},
ticks=none,
axis x line*=bottom,
axis y line*=left
]
\end{axis}
\end{tikzpicture}%

}
\end{subfigure}
\caption{Statistics of the user responses in study 1. (Top) The \textit{algorithm} factor. (Bottom) The \textit{segment} factor.}
\label{fig:exp 2 means}
\end{figure}

We manipulated two independent factors: the \textit{algorithm} (with 4 levels: DDM, DPP, GD, and baseline) and the \textit{segment} at which the trajectory is evaluated (with 4 levels: 25\%, 50\%, 75\%, and 90\% of the total length, shown in Fig. \ref{simulation_tikz2} (left)), leading to a total of 16 conditions. 
We used two dependent variables to measure deceptiveness: (i) \textit{goal prediction incorrectness} and (ii) \textit{incorrect prediction confidence}. We used a between-subjects design and recruited 320 users (20 per condition) on Amazon’s Mechanical Turk. For each condition, we showed users the corresponding trajectory segment and asked them (i) to predict the agent's goal and (ii) to state their confidence on a 5-point Likert scale.

A factorial ANOVA on \textit{goal prediction incorrectness} (considered to be robust to dichotomous data \cite{d1971second}) revealed significant main effects for both \textit{algorithm} ($F(2,228)$$=$$27.344$, $p$$<$$0.001$) and \textit{segment} ($F(3,228)$$=$$59.817$, $p$$<$$0.001$) as well as significant interaction effects ($F(6,228)$$=$$4.949$, $p$$<$$0.001$). A factorial ANOVA on \textit{incorrect prediction confidence} revealed similar significant main and interaction effects.

In line with $\textbf{H}_1$, a post-hoc analysis with Tukey HSD that marginalizes over segments showed that DDM and DPP  are significantly more deceptive than GD and baseline ($p$$<$$0.001$ for all pairwise comparisons). Fig. \ref{fig:exp 2 means} echoes these findings, where we  
plot the means and the standard errors of the dependent variables. Note that DDM induces wrong goal predictions 10 times more often than the baseline and 2 times more often than GD. Moreover, both DDM and DPP induce wrong predictions up to $50\%$ segment of the trajectories, whereas GD  reveals the true goal with high confidence after $25\%$ segment of the trajectory.

\subsubsection{Study 2: the importance of prediction-awareness} Next, we consider the environment shown in Fig. \ref{simulation_tikz2} (right) which includes 4 potential goals \textbf{G1},\dots,\textbf{G4}. The true goal is \textbf{G1}. 

Recall that the DDM algorithm systematically generates exaggerated trajectories using prediction probabilities. In complex environments with multiple decoy goals and obstacles, we expect   DDM to be more deceptive than heuristic approaches, e.g., DPP. We also expect the GD algorithm's local optimality to limit its deceptiveness in this complex environment. The trajectory generated by the DPP algorithm first pretends to reach the decoy goal \textbf{G3}, which is the closest decoy to the true goal. Hence, DDM coincides with the baseline. In this study, we hypothesize the following.

$\textbf{\textup{H}}_2$: \textit{DDM generates significantly more deceptive trajectories than DPP and GD.}  
\input{exp_3_tikz}

We manipulated two independent factors: the \textit{algorithm} (with 3 levels: DDM, DPP, and GD) and the \textit{segment} at which the trajectory is evaluated (with 4 levels shown in Fig. \ref{simulation_tikz2} (right)), leading to a total of 12 conditions. We used a between-subjects design and recruited 240 users (20 per condition) on Amazon’s Mechanical Turk. To measure deceptiveness, we used the two dependent variables from the previous study and \textit{two-goal prediction incorrectness}. For each condition, we asked users (i) to predict the agent's goal, (ii) to state their confidence on a 5-point Likert scale, and (iii) to predict the agent's second most likely goal.  A two-goal prediction is incorrect if the goal prediction \textit{and} the second most likely goal prediction are different than the true goal. Note that we asked the users second most likely goal to understand the effect of the decoy goal \textbf{G3} on predictions.

A factorial ANOVA analysis yielded significant main and interaction effects for all dependent variables.
In line with $\textbf{H}_2$, a post-hoc analysis with Tukey HSD that marginalizes over segments revealed that DDM is significantly more deceptive than DPP with respect to all three dependent variables ($p$$<$$0.001$ for all comparisons). There is no significant difference between the DDM and GD with respect to \textit{goal prediction incorrectness} and \textit{incorrect prediction confidence} variables. However, the comparison with respect to \textit{two-goal prediction incorrectness} variable revealed that DDM is significantly more deceptive than GD ($p$$<$$0.001$). The means and standard errors depicted in Fig. \ref{fig:exp 3 means} also reflect these findings. Note in the figure that the deceptiveness of the DDM only slightly changes when the users' second most likely goal prediction is included in the analysis, whereas the deceptiveness of GD and DPP dramatically decreases.


\subsection{Deception Under Probabilistic Constraints}
\begin{figure}[t!]
\centering
\begin{tikzpicture}
    \node[inner sep=0] at (0,0) {\includegraphics[width=0.3\textwidth]{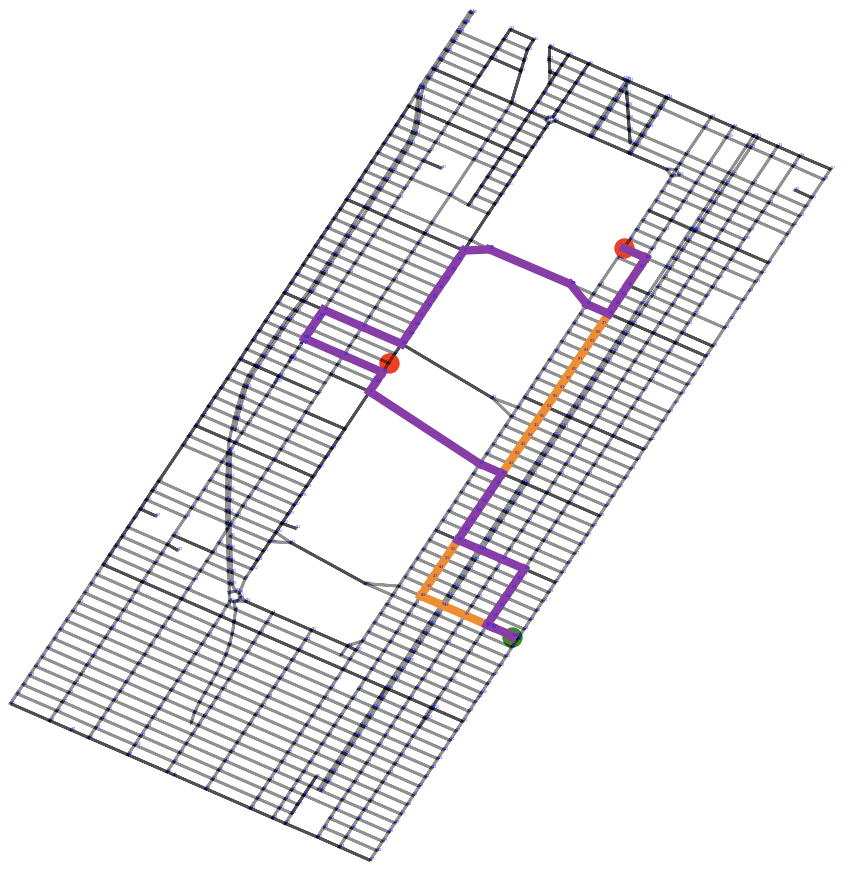}};

    \node[opacity = 1] at (1.8,-1.25) {\scriptsize initial state};
    \draw[-latex, line width=0.2mm, color = black, opacity = 0.6] (0.7,-1.25) -- (1.3,-1.25);
    
    \node[] at (1,1.65) {\tiny true goal};
    \draw[-latex, line width=0.2mm, color = black, opacity = 0.6] (1.2,1.28) -- (1,1.55);
    
    \node[] at (0.4,0.8) {\tiny decoy};
    \node[] at (0.4,0.6) {\tiny goal};
    \draw[-latex, line width=0.2mm, color = black, opacity = 0.6] (-0.25,0.45) -- (0.2,0.7);
    
    \draw [draw=violet,line width = 0.05 cm] (-3.55,1.65) rectangle (-0.55,2.25);
    \node[] at (-2.05,2.05) {\tiny reach the true goal in \textbf{40 minutes}};
    \node[] at (-2.3,1.85) {\tiny with at least 0.8 probability};
    \draw[-latex, line width=0.2mm, color = black, opacity = 0.6] (-0.7,0.9) -- (-1.6,1.55);
  
    \draw [draw=orange,line width = 0.05 cm] (1.3,-0.4) rectangle (4.3,-1);  
    \node[] at (2.8,-0.6) {\tiny reach the true goal in \textbf{30 minutes}};
    \node[] at (2.55,-0.8) {\tiny with at least 0.8 probability};
    \draw[-latex, line width=0.2mm, color = black, opacity = 0.6] (0.9,0.3) -- (1.8,-0.35);
    
\end{tikzpicture}
\caption{An illustration of deceptive trajectories in Manhattan case study. The agent exaggerates its behavior by moving towards the decoy goal only when the probabilistic constraint on arrival time allows such a behavior.}
\label{manhattan}
\end{figure}

We now consider a large-scale example and demonstrate how the proposed algorithm can generate deceptive trajectories while respecting probabilistic constraints on travel time. 

We consider the graph given in Fig. \ref{manhattan}, which represents the road network in Manhattan, New York. We utilize the real-world speed data provided in the open source database \cite{uber} to express realistic travel times. We generate a continuous travel time distribution on each edge by assuming
that the speed follows a lognormal distribution, which is a common assumption in transportation networks \cite{rakha2010trip}. To construct the MDP model expressing stochastic travel times, we discretize the travel time distributions and take the Cartesian product of the graph with the set $\{0.5, 1, 1.5,\dots, T_{\max}\}$ of states, where $T_{\max}$ is the maximum travel time in minutes. In this MDP, the agent's transition from a state $(s,t)$ to $(s',t')$ with probability $p$ expresses that the agent's travel from $s$ to $s'$ takes $t'$$-$$t$ minutes with probability $p$.

We consider two potential goals shown in Fig. \ref{manhattan} and synthesize two exaggerated trajectories ensuring that the agent reaches its true goal in $T_{\max}$$\in$$\{30,40\}$ minutes with 0.8 probability. Note that one can encode this constraint in the proposed framework by slightly changing the constraint in \eqref{deceptive_cons} and defining it as $\text{Pr}(Reach[(G^{\star},T_{\max})])$$\geq$$0.8$ on the constructed MDP. We choose the value $0.8$ to more clearly illustrate the effect of probabilistic time constraints on the agent's deceptive behavior. Finally, to synthesize the optimal deceptive policies, we use the parameters $c(s,a)$$=$$5$ for all $s$$\in$$\mathcal{S}$ and $a$$\in$$\mathcal{A}$, $\gamma_o$$=$$0.95$, $\alpha$$=$$1$, and $\gamma_a$$=$$1$.

The two trajectories shown in Fig. \ref{manhattan} demonstrates that the proposed algorithm enables the agent to adjust its exaggeration behavior with respect to the desired travel time. As can be seen from the figure, when the agent is required to arrive its goal in 30 minutes with at least 0.8 probability, it simply follows a shortest trajectory to the goal. This is because the agent's stochastic travel time constraint prevents it from exaggerating its behavior. Indeed, when the agent is required to arrive its goal in 40 minutes instead of 30, its trajectory crosses to the left side of the road network and pretends to reach the decoy goal before reaching the true goal.

\section{Conclusions}

We consider an autonomous agent that aims to reach one of multiple potential goals in a stochastic environment and propose a novel approach to generate globally optimal deceptive strategies via linear programming. We evaluate the performance of the proposed approach via comparative user studies and present a case study on the streets of Manhattan, New York illustrating the use of deception in realistic scenarios under probabilistic constraints. Future work will focus on characterizing the sensitivity of the deceptive strategies to the knowledge of the observer's prediction model.

\newpage
\bibliography{bib_aaai}
\bibliographystyle{aaai}

\end{document}


\title{Deceptive Decision-Making Under Uncertainty: Technical Appendix}
\author{Anonymous Author(s) \\ ~ \\ Paper ID: 10696}
\maketitle
In this technical appendix, we provide additional information on the experimental results presented in the main body of the paper to ensure reproducibility.

\subsection{User Studies}
In the user studies, we compare the performance of the proposed algorithm, i.e., DDM, with two existing algorithms, i.e., GD \cite{dragan2015deceptive} and DPP \cite{masters2017deceptive}, as well as a baseline. In what follows, we provide a detailed list of the parameters used in each algorithm.
\subsubsection{Study 1: the importance of global optimality} The following are the parameters used in the first user study. Note that we also provide the code to generate the trajectories, the raw data collected from Amazon Mechanical Turk, and the statistical analyses conducted using the IBM SPSS Statistics \cite{spss} as a supplementary material.

\textbf{Baseline:} We generate the baseline trajectory by choosing $\gamma_a$$=$$\gamma_o$$=$$0.95$, $c(s,a)$$=$$1$ for each $s$$\in$$\mathcal{S}$ and $a$$\in$$\mathcal{A}$, $\alpha$$=$$20$, and $P(G)$$=$$1/2$ for each $G$$\in$$\mathcal{G}$. To terminate the softmax value iteration, we use the condition $\sum_{G\in\mathcal{G}}\lvert V^t_G(s) - V^t_G(s)\rvert$$\leq$$10^{-4}$ where $V^{t}_G(s)$ is the value at the $t$-th time step. To avoid numerical instabilities, we also impose the condition that $g(s,a)$$=$$0$ if $g(s,a)$$\leq$$10^{-5}$.

\textbf{DDM:} We generate the DDM trajectory by choosing $\gamma_a$$=$$\gamma_o$$=$$0.95$, $c(s,a)$$=$$1$ for each $s$$\in$$\mathcal{S}$ and $a$$\in$$\mathcal{A}$, $\alpha$$=$$0.5$, and $P(G)$$=$$1/2$ for each $G$$\in$$\mathcal{G}$. To terminate the softmax value iteration, we use the condition $\sum_{G\in\mathcal{G}}\lvert V^t_G(s) - V^t_G(s)\rvert$$\leq$$10^{-4}$ where $V^{t}_G(s)$ is the value at the $t$-th time step. To avoid numerical instabilities, we also impose the condition that $g(s,a)$$=$$0$ if $g(s,a)$$\leq$$10^{-5}$.

\textbf{GD:} We parameterize the GD trajectories as a vector of $40$ waypoints. To ensure obstacle avoidance we use the potential function $\phi$$:$$\mathcal{X}$$\rightarrow$$\mathbb{R}$ where $\mathcal{X}$ denotes the continuous domain of the $9$$\times$$9$ grid world. The function $\phi$ is defined as $\phi(x)$$=$$1$ if $d_x$$\leq$$0.16$, $\phi(x)$$=$$(\cos(\pi\frac{d_x-0.16}{0.8})+1)/2$ if $0.16$$\leq$$d_x$$\leq$$0.8$ and $\phi(x)$$=$$0$ if $d_x$$>$$0.8$, where $d_x$ is the closest distance of the point $x$$\in$$\mathcal{X}$ to an obstacle in the environment. We set the constraint budget to $\beta$$=$$6$ (see \cite{Draganlegible} for a discussion on the constraint budget.) Finally, for the gradient descent algorithm, we set the learning rate to $1/\eta$$=$$1/100$ and terminate the algorithm when either the gradient is less than equal to $10^{-3}$ or the iteration number exceeds 300.

\textbf{DPP:} For the DPP algorithm, we obtained the trajectory by computing the shortest trajectory to the decoy goal from the initial state and the shortest trajectory from the decoy goal to the true goal. For implementation, we utilized the open source code provided by the authors. The authors original implementation allows the agent to move diagonally in the environment. Since the agent cannot move diagonally in the examples considered in the paper, we eliminated such diagonal actions by assigning a cost of 100 to them.

\subsubsection{Study 2: the importance of prediction-awareness} The following are the parameters used in the second user study. Note that we also provide the code to generate the trajectories, the raw data collected from Amazon Mechanical Turk, and the statistical analyses conducted using the IBM SPSS Statistics \cite{spss} as a supplementary material.

\textbf{Baseline:} We use the same parameters as in the first study. Note that DPP and the baseline generate the same trajectories in this study.

\textbf{DDM:}We generate the DDM trajectory by choosing $\gamma_a$$=$$0.9$ $\gamma_o$$=$$0.95$, $c(s,a)$$=$$20$ for each $s$$\in$$\mathcal{S}$ and $a$$\in$$\mathcal{A}$, $\alpha$$=$$1$, and $P(G)$$=$$1/2$ for each $G$$\in$$\mathcal{G}$. To terminate the softmax value iteration, we use the condition $\sum_{G\in\mathcal{G}}\lvert V^t_G(s) - V^t_G(s)\rvert$$\leq$$10^{-4}$ where $V^{t}_G(s)$ is the value at the $t$-th time step. To avoid numerical instabilities, we also impose the condition that $g(s,a)$$=$$0$ if $g(s,a)$$\leq$$10^{-5}$.

\textbf{GD:} We use the same parameters as in the first study.

\textbf{DPP:} We use the same parameters as in the first study.

\subsection{Deception Under Probabilistic Constraints}
Note that we provide the code for this study, as well as the travel time data extracted from \cite{uber}, as a supplementary material.

In this case study, we first construct the MDP corresponding to the map of Manhattan, New York using the travel time data extracted from \cite{uber}. Specifically, we form the MDP $\mathcal{M}$$=$$(\mathcal{S},s_1, \mathcal{A},P)$ where each $s$$\in$$\mathcal{S}$ corresponds to an intersection point of multiple roads, $s_1$ is the agent's initial location, $\mathcal{A}$ is the number of neighboring intersections the agent can transition to, and $P$ is the deterministic transition function expressing the connected roads in the network.

We obtain the travel time distribution between each permissible state pair $(s,s')$ from the speed data (provided as mean and variance) by assuming that the speed follows a lognormal distribution. We discretize the continuous lognormal distribution using $30$ seconds time buckets. Specifically, let $T_{\max}$ be the maximum travel time in minutes. We discretize the distribution into $2T_{\max}$$+$$1$ intervals. By doing so we express the probability $F((s,s'), T_1\leq t\leq T_2)$ that the agent's travel time $t$ over the edge $(s,s')$ is between $T_1$ and $T_2$ seconds using the corresponding interval. Let $k$ be the random travel time over the edge $(s,s')$. Note that, in this formulation, we have $\sum_{k=1}^{2T_{\max}+1}F((s,s'),k)$$=$$1$ for each permissible $(s,s')$ pair. 

From the MDP $\mathcal{M}$ and the discretized travel time distributions, we obtain the product MDP as follows. Let $\mathcal{S}$$\times$$\{1,2,\ldots,2T_{\max}+1\}$ be the states in the product model. We define the transition function $P'$ as 
\begin{align*}
&P'((s,t),a,(s',t'))=\\
&\begin{cases}
F((s,s'),t'-t) & \ \text{if} \ t'\leq T_{\max}\\
\sum_{k : t+k > T_{\max}} F((s,s'),k) & \ \text{if} \ t' = 2T_{\max}+1\\
0 & \ \text{otherwise}.
\end{cases}
\end{align*}
Intuitively, if the agent takes the action $a$ to transition from $(s,t)$ to $(s',t')$, it transitions to $(s',t')$ with probability $F((s,s'),t'-t)$ since the travel time over the edge $(s,s')$ is $t'$$-$$t$ with probability $F((s,s'),t'-t)$. Moreover, if there is a nonzero probability that the agent's total travel time exceeds the budget $T_{\max}$, then the agent transitions to the state $(s',2T_{\max}+1)$ with the corresponding probability.

On the product MDP, the probability of reaching the true goal $G^{\star}$ within $T_{\max}$ minutes corresponds to the total reachability probability of the states $(G^{\star},t)$ where $t$$\leq$$2T_{\max}$. Hence, to satisfy the desired the probabilistic reachability condition, we use the constraint $\sum_{k=1}^{2T_{\max}}\text{Pr}(Reach[(G^{\star},k)])$$\geq$$0.8$ on the product model. 

Finally, we set the costs $g((s,t),a)$$=$$g(s,a)$ on the product model and use the proposed approach presented in the main body of the paper to synthesize the deceptive policies.
\bibliography{bib_aaai}
\bibliographystyle{aaai}